%% file: sample-sigconf-authordraft.tex
\documentclass[sigconf,authorversion]{acmart}

\usepackage{booktabs} 

\usepackage[utf8]{inputenc} 
\usepackage[T1]{fontenc}    
\usepackage{url}            
\usepackage{booktabs}       
\usepackage{amsfonts}       
\usepackage{nicefrac}       
\usepackage{microtype}      
\usepackage{graphicx}
\usepackage{subcaption}
\usepackage[title]{appendix}
\usepackage{verbatim}
\usepackage{EvolvingAIcommenting}

\setcopyright{rightsretained}

\copyrightyear{2018} 
\acmYear{2018} 
\acmConference[GECCO '18]{Genetic and Evolutionary Computation Conference}{July 15--19, 2018}{Kyoto, Japan}
\acmBooktitle{GECCO '18: Genetic and Evolutionary Computation Conference, July 15--19, 2018, Kyoto, Japan}
\acmPrice{15.00}
\acmDOI{10.1145/3205455.3205474}
\acmISBN{978-1-4503-5618-3/18/07}

\begin{document}
\title{ES Is More Than Just a Traditional Finite-Difference Approximator}

\author{Joel Lehman, Jay Chen, Jeff Clune, and Kenneth O. Stanley}
\affiliation{%
  \institution{Uber AI Labs, San Francisco, CA}
}
\email{ {joel.lehman,jayc,jeffclune,kstanley}@uber.com}

\begin{abstract}
An evolution strategy (ES) variant based on a simplification of a natural evolution strategy 
recently attracted attention because it performs surprisingly well in challenging deep reinforcement learning domains. It searches for neural network parameters by generating perturbations to the current set of parameters, checking their performance, and moving in the aggregate direction of higher reward. Because it resembles a traditional finite-difference approximation of the reward gradient, it can naturally be confused with one. However, this ES optimizes for a different gradient than just reward: It optimizes for
the average reward of the entire population, thereby seeking parameters that are robust to perturbation. This difference can channel ES into distinct areas of the search space relative to gradient descent, and also consequently to networks with distinct properties.  This unique robustness-seeking property, and its consequences for optimization, are demonstrated in several domains. They include humanoid locomotion, where networks from policy gradient-based reinforcement learning are significantly less robust to parameter perturbation than ES-based policies solving the same task.  While the implications of such robustness and robustness-seeking remain open to further study, this work's main contribution is to highlight such differences and their potential importance.
\end{abstract}

%
%


\maketitle

\input{samplebody-conf}

\input{si}

\end{document}

%% file: samplebody-conf.tex




\section{Introduction}




\citet{salimans:es} recently demonstrated that an approach they call an \emph{evolution strategy} (ES) can compete on modern reinforcement learning (RL) benchmarks that require large-scale deep learning architectures. 
While ES is a research area with a rich history \citep{schwefel:es} 
encompassing a broad variety of search algorithms (see \citet{beyer:nc02}), \citet{salimans:es} has drawn attention to the particular form of ES applied in that paper (which does not reflect the field as a whole), in effect a simplified version of natural ES (NES; \citep{wierstra:jmlr14}). Because this form of ES is the focus of this paper, herein it is referred to simply as \emph{ES}. One way to view ES is as a policy gradient algorithm applied to the \emph{parameter space} instead of to the \emph{state space} as is more typical in RL \citep{williams:reinforce}, and the distribution of parameters (rather than actions) is optimized to maximize the expectation of performance. Central to this interpretation is how ES estimates (and follows) the gradient of increasing performance with respect to the current distribution of parameters. In particular, in ES many independent parameter vectors are drawn from the current distribution, their performance is evaluated, and this information is then aggregated to estimate a gradient of distributional improvement. 

The implementation of this approach bears similarity to a finite-differences (FD) gradient estimator \citep{richardson:fd,spall:spsa}, wherein evaluations of \emph{tiny} parameter perturbations are aggregated into an estimate of the performance gradient. As a result, from a non-evolutionary perspective it may be attractive to interpret the results of \citet{salimans:es} solely through the lens of FD (e.g.\ as in \citet{ebrahimi:arxiv17}), concluding that the method is interesting or effective only because it is approximating the gradient of performance with respect to the parameters. However, such a hypothesis ignores that ES's objective function is interestingly different from traditional FD, which this paper argues grants it additional properties of interest. In particular, ES optimizes the performance of \emph{any draw} from the learned \emph{distribution} of parameters (called the search distribution), while FD optimizes the performance of \emph{one particular} setting of the domain parameters. The main contribution of this paper is to support the hypothesis that this subtle distinction may in fact be important to understanding the behavior of ES (and future NES-like approaches), by conducting experiments that highlight how ES is driven to more \emph{robust} areas of the search space than either FD or a more traditional evolutionary approach. The push towards robustness carries potential implications for RL and other applications of deep learning that could be missed without highlighting it specifically.

Note that this paper aims to clarify a subtle but interesting possible misconception, not to debate what exactly qualifies as a FD approximator. The framing here is that a traditional finite-difference gradient approximator makes tiny perturbations of domain parameters to estimate the gradient of improvement for the current \emph{point} in the search space. While ES also stochastically follows a gradient (i.e.\ the \emph{search gradient} of how to improve expected performance across the search distribution representing a \emph{cloud} in the parameter space), it does not do so through common FD methods. In any case, the most important distinction is that ES optimizes the expected value of a distribution of parameters with fixed variance, while traditional finite differences optimizes a singular parameter vector. 

To highlight systematic empirical differences between ES and FD, this paper first uses simple two-dimensional fitness landscapes. These results are then validated in the Humanoid Locomotion RL benchmark domain, showing that ES's drive towards robustness manifests also in complex domains: Indeed, parameter vectors resulting from ES are much more robust than those of similar performance discovered by a genetic algorithm (GA) or by a non-evolutionary policy gradient approach (TRPO) popular in deep RL. These results have implications for researchers in evolutionary computation (EC; \citep{dejong:book02}) who have long been interested in properties like robustness \citep{wilke:nature01,wagner:robustness,lenski:balancing} and evolvability \citep{wagner:evolvability,kirschner:evolvability,lehman:plos13}, and also for deep learning researchers seeking to more fully understand ES and how it relates to gradient-based methods. 

\vspace{-0.05in}
\section{Background}


This section reviews FD, the concept of search gradients (used by ES), and the general topic of robustness in EC.

\vspace{-0.05in}
\subsection{Finite Differences}


A standard numerical approach for estimating a function's gradient is the \emph{finite-difference} method. In FD, tiny (but finite) perturbations are applied to the parameters of a system outputting a scalar. Evaluating the effect of such perturbations
enables approximating the derivative with 
respect to the parameters. Such a method is useful for optimization when the system is not differentiable, e.g.\ in  RL, when reward comes from a partially-observable or analytically-intractable environment. Indeed, because of its simplicity there are many 
policy gradient methods motivated by FD \citep{spall:spsa,glynn:likelilood}.

One common finite-difference estimator of the derivative of function $f$ with respect to the scalar $x$ is given by: \vspace{-0.05in}
$$ f'(x) \approx \frac{f(x+\epsilon)-f(x)}{\epsilon},$$
given some small constant $\epsilon$. This estimator generalizes naturally to \emph{vectors} of parameters, where the partial derivative
with respect to each vector element can be similarly calculated; however, naive FD
scales poorly to large parameter vectors, as it perturbs each parameter \emph{individually}, making its
application infeasible for large problems (like optimizing deep neural networks). However, FD-based
methods such as
simultaneous 
perturbation stochastic approximation (SPSA; \citealt{spall:spsa}) can aggregate information from independent perturbations of
all parameters simultaneously to estimate the gradient more efficiently. Indeed, SPSA is similar in implementation to ES.

However, the theory for FD methods relies on \emph{tiny} perturbations; the larger such perturbations become, the less meaningfully FD approximates the underlying gradient, which formally is the slope of the function with respect to its parameters
at a particular point. In other words, as perturbations become larger, FD becomes qualitatively disconnected from the principles
motivating its construction; its estimate becomes increasingly influenced by the curvature of the reward function, and its interpretation becomes unclear. 
This consideration is important because ES is \emph{not} motivated by tiny perturbations nor by approximating the
gradient of performance for any singular setting of parameters, as described in the next section.

\vspace{-0.05in}
\subsection{Search Gradients}

Instead of searching directly for one high-performing parameter vector, as is typical in gradient descent and FD
methods, a distinct approach is to optimize the \emph{search distribution} of domain parameters to achieve high average reward when
a particular parameter vector is sampled from the distribution \citep{berny:searchgradient,wierstra:jmlr14,sehnke:pgpe}. Doing so
requires following \emph{search gradients} \citep{berny:searchgradient,wierstra:jmlr14}, i.e.\ gradients of increasing expected
fitness with respect to distributional parameters (e.g.\ the mean and variance of a Gaussian distribution).

While the procedure for following such search gradients uses mechanisms similar to a FD gradient approximation (i.e.\ it
involves aggregating fitness information from samples of domain parameters in a local neighborhood),
importantly the underlying objective function from which it derives is different:

\vspace{-.1in}
\begin{equation}
J(\theta) = \mathrm{E}_\theta f(z) = \int f(z) \pi(z|\theta) dz,
\label{eq:searchgrad}
\end{equation}
where $f(z)$ is the fitness function, and $z$ is a sample from the search distribution $\pi(z|\theta)$ specified by parameters $\theta$. Equation
\ref{eq:searchgrad} formalizes the idea that ES's objective (like other search-gradient methods) is to optimize the \emph{distributional parameters} such that the expected 
fitness of \emph{domain parameters} drawn from that search distribution is maximized. In contrast, the objective function for more traditional
gradient descent approaches is to find the optimal domain parameters directly: $J(\theta) = f(\theta)$.

While NES allows for adjusting both the mean and variance of a search distribution, in the ES of \citet{salimans:es}, the evolved
distributional parameters control only the \emph{mean} of a Gaussian distribution and not its variance. As a result, ES cannot reduce
variance of potentially-sensitive parameters; importantly, the implication
is that ES will be driven towards \emph{robust} areas of the search space. For example, imagine two paths through the search space of similarly increasing
reward, where one path requires precise settings of domain parameters (i.e.\ only a low-variance search distribution could capture such precision) while the other does not. 
In this scenario, ES with a sufficiently-high variance setting will only be able to follow
the latter path, in which performance is generally robust to parameter perturbations. The experiments in this paper illuminate circumstances in which this robustness property of ES impacts search. Note that the relationship of \emph{low-variance} ES (which bears stronger similarity to finite differences) to stochastic gradient descent (SGD) is explored in more depth in \citet{zhang:arxiv17}.

\subsection{Robustness in Evolutionary Computation}

Researchers in EC have long been concerned with robustness in the face of mutation \citep{wilke:nature01,wagner:robustness,lenski:balancing}, i.e.\ the idea that randomly
mutating a genotype will not devastate its functionality. In particular, evolved genotypes
in EC often \emph{lack} the apparent robustness of natural organisms \citep{lehman:selfadapt}, which can hinder progress in an evolutionary algorithm (EA). In other words, robustness is important for its
link to \emph{evolvability} \citep{kirschner:evolvability,wagner:evolvability}, or the ability of evolution to
generate productive heritable variation. 

As a result, EC researchers have introduced mechanisms useful to encouraging robustness, such as
self-adaptation \citep{meyer:selfadapt}, wherein evolution can modify or control aspects of generating
variation. Notably, however, such mechanisms can emphasize robustness \emph{over} evolvability depending on selection pressure \citep{clune:selfadapt,lehman:selfadapt}, i.e.\ robustness can be trivially maximized when
a genotype encodes that it should be subjected only to trivial perturbations. ES avoids this potential 
pathology because the variance of its distribution is fixed, although in a full implementation of NES 
variance is subject to optimization and the robustness-evolvability trade-off would likely re-emerge.

While the experiments in this paper show that ES is drawn to robust areas of the search space
as a direct consequence of its objective (i.e.\ to maximize expected fitness across its search
distribution), in more traditional EAs healthy robustness is often a second-order effect \citep{lehman:selfadapt,kounios:evolvability,wilke:nature01}. For example, if
an EA lacks elitism and mutation rates are high, evolution favors more robust optima although it is not a direct objective of search \citep{wilke:nature01}; similarly, when selection pressure rewards phenotypic or behavioral divergence, self-adaptation can serve to balance robustness and evolvability \citep{lehman:selfadapt}. 

Importantly, the relationship between ES's robustness drive and evolvability is nuanced and likely domain-dependent. For example, some domains
may indeed require certain NN weights to be precisely specified, and evolvability may be hindered by prohibiting such specificity. Thus
an interesting open question is whether ES's mechanism for generating robustness can be enhanced to better seek evolvability in a domain-independent way, and additionally, whether its mechanism can be
abstracted such that its direct search for robustness can also benefit more traditional EAs.

\vspace{-0.05in}
\section{Experiments}


%


This section empirically investigates how the ES of \citet{salimans:es} systematically differs from more traditional gradient-following approaches. First, through a series of toy landscapes, a FD approximator of domain parameter improvement is contrasted with ES's approximator of distributional parameter improvement. 
Then, to ground such findings, the robustness property of ES is further investigated in a popular RL benchmark, i.e.\ the Humanoid Locomotion task \citep{brockman}. Policies
from ES are compared with those from a genetic algorithm (GA) and a representative high-performing policy gradient algorithm (TRPO; \citep{schulman:trpo}), to explore whether ES is drawn to qualitatively different areas of the parameter space.



\vspace{-0.05in}
\subsection{Fitness Landscapes}

\def \mszz {1.3in}
\def \msz {1.85in}
\def \cutdist {-0.2in}
\def \msbig {1.8in}
\def \cutdistbig {-0.4in}

This section introduces a series of illustrative fitness landscapes (shown in figure \ref{fig:landscape}) wherein the behavior of ES and FD can easily be contrasted. In each landscape, performance is
a deterministic function of two variables. For ES, the distribution over variables is an isotropic Gaussian with fixed variance as in \citet{salimans:es}; i.e.\ ES optimizes two distributional parameters that encode the location of the distribution's \emph{mean}. In contrast, while the FD gradient-follower also optimizes two parameters, these represent a \emph{single} instantiation of domain parameters, and consequently its function thus depends only on $f(\theta)$ at that singular position. The FD algorithm applies a central difference gradient estimate for each parameter independently. Note that unless otherwise specified, the update rule for the gradient-follower is vanilla gradient descent (i.e.\ it has a fixed learning rate and no momentum term); one later fitness landscape experiment will explore whether qualitative differences between ES and FD can be bridged through combining finite differences with more sophisticated optimization heuristics (i.e.\ by adding momentum).

\def \mszz {1.2in}
\begin{figure*}[h]
  \centering
  \begin{subfigure}[t]{0.19\textwidth}
  		\centering
         \includegraphics[height=\mszz]{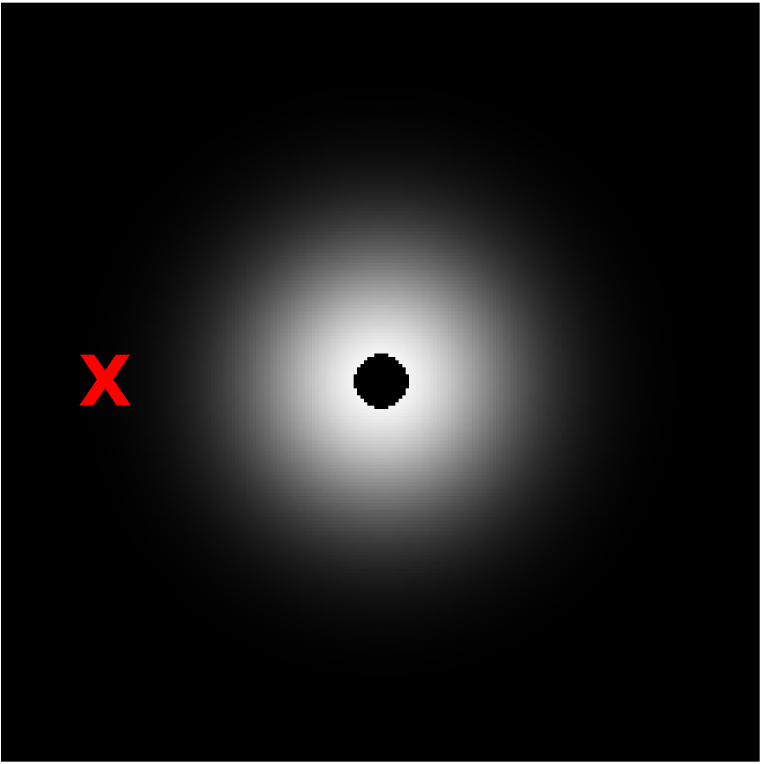}
        \caption{Donut}
    \end{subfigure}
  \begin{subfigure}[t]{0.19\textwidth}
        \centering
  \includegraphics[height=\mszz]{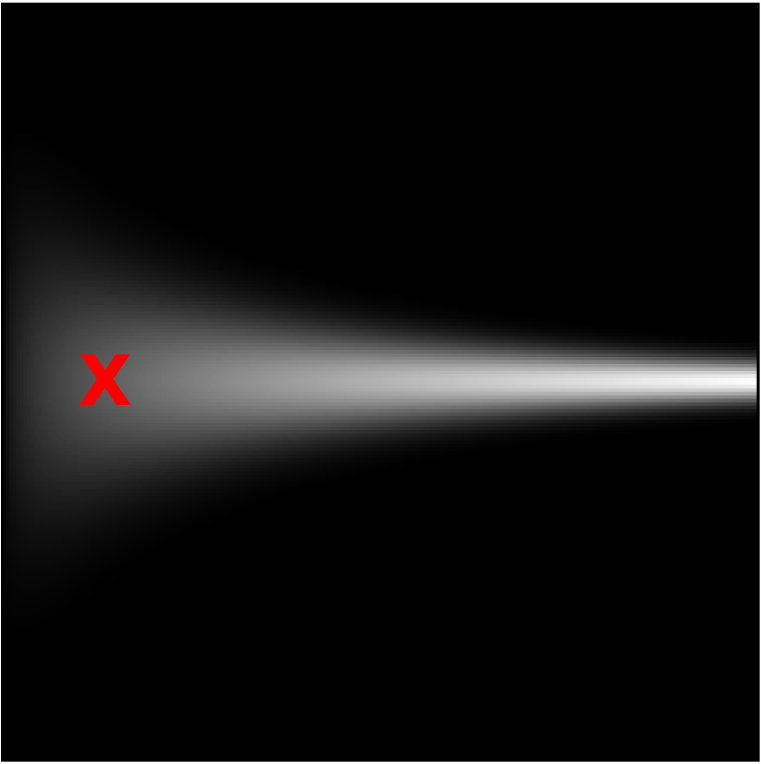}
  \caption{Narrowing Path}
  \end{subfigure}
 \begin{subfigure}[t]{0.19\textwidth}
        \centering
  \includegraphics[height=\mszz]{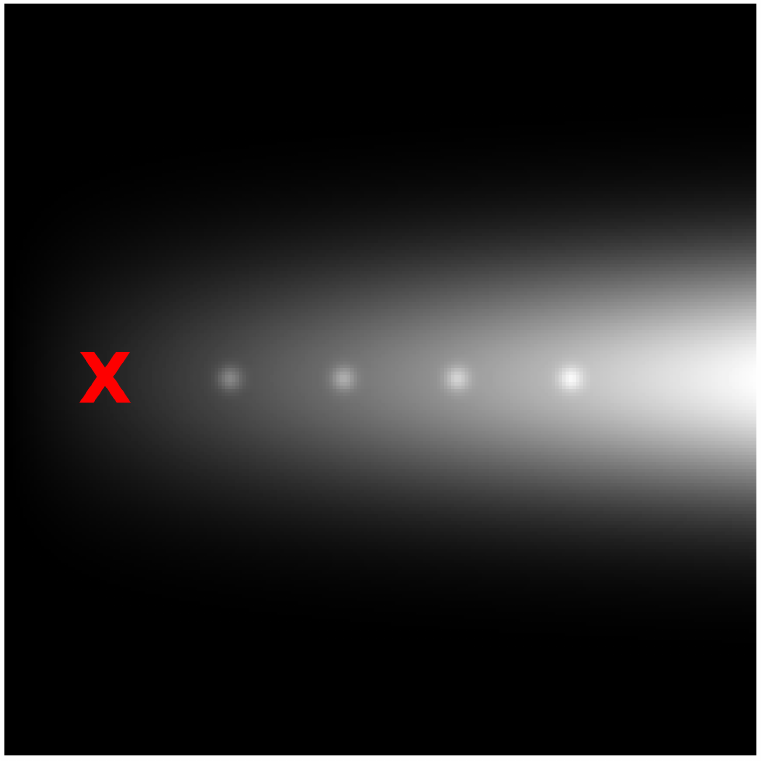}
  \caption{Fleeting Peaks} 
  \end{subfigure} 
  \begin{subfigure}[t]{0.19\textwidth}
  		\centering
         \includegraphics[height=\mszz]{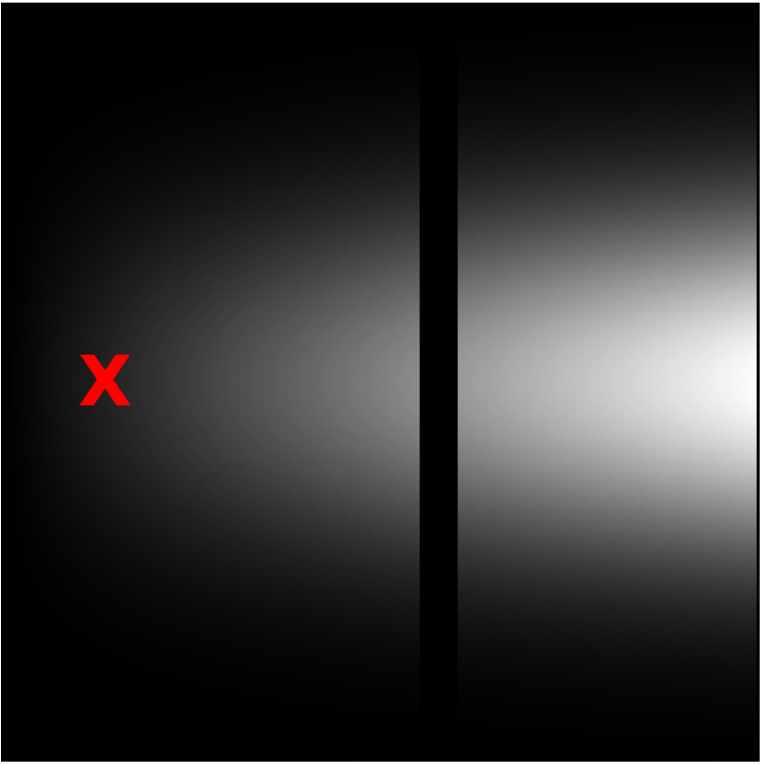}
        \caption{Gradient Gap}
    \end{subfigure}
  \begin{subfigure}[t]{0.19\textwidth}
        \centering
  \includegraphics[height=\mszz]{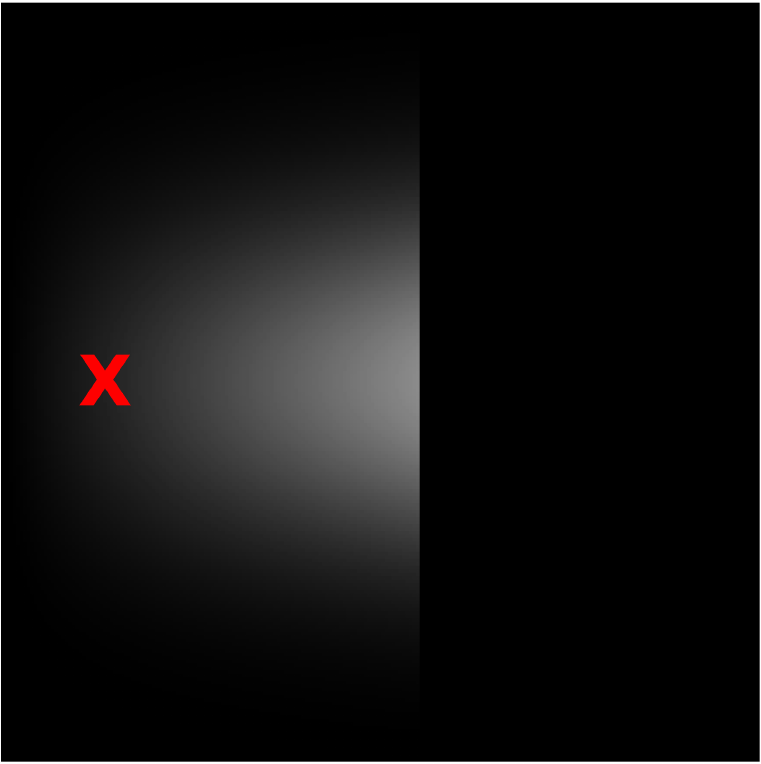}
  \caption{Gradient Cliff}
  \end{subfigure}
	\vspace{-0.1in}
  \caption{\textbf{Illustrative fitness landscapes.} A series of five fitness landscapes highlight divergences between the behavior of ES and FD. In all landscapes, darker colors indicate lower fitness and the red X indicates the starting point of search. In the (a) Donut landscape, a Gaussian function assigns fitness to each point, but the small neighborhood immediately around and including the Gaussian's peak is flattened to a reward of zero. In the (b) Narrowing Path landscape, fitness increases to the right, but the peak's spread increasingly narrows, testing an optimizer's ability to follow a narrow path. In the (c) Fleeting Peaks landscape, fitness increases to the right, but optimization to the true peak is complicated by a series of small local optima. The (d) Gradient Gap landscape is complicated by a gradient-free zero-reward gap in an otherwise smooth landscape, highlighting ES's ability to cross fitness plateaus (i.e.\ escape areas of the landscape where there is no local gradient). A control for the Gradient Gap landscape is the (e) Gradient Cliff landscape, wherein there is no promising area beyond the gap.\label{fig:landscape}}
    
\end{figure*}

\begin{figure}[h]
	\vspace{-0.2in}
  \centering
  \begin{subfigure}[t]{0.24\textwidth}
  		\centering
         \includegraphics[height=\msbig]{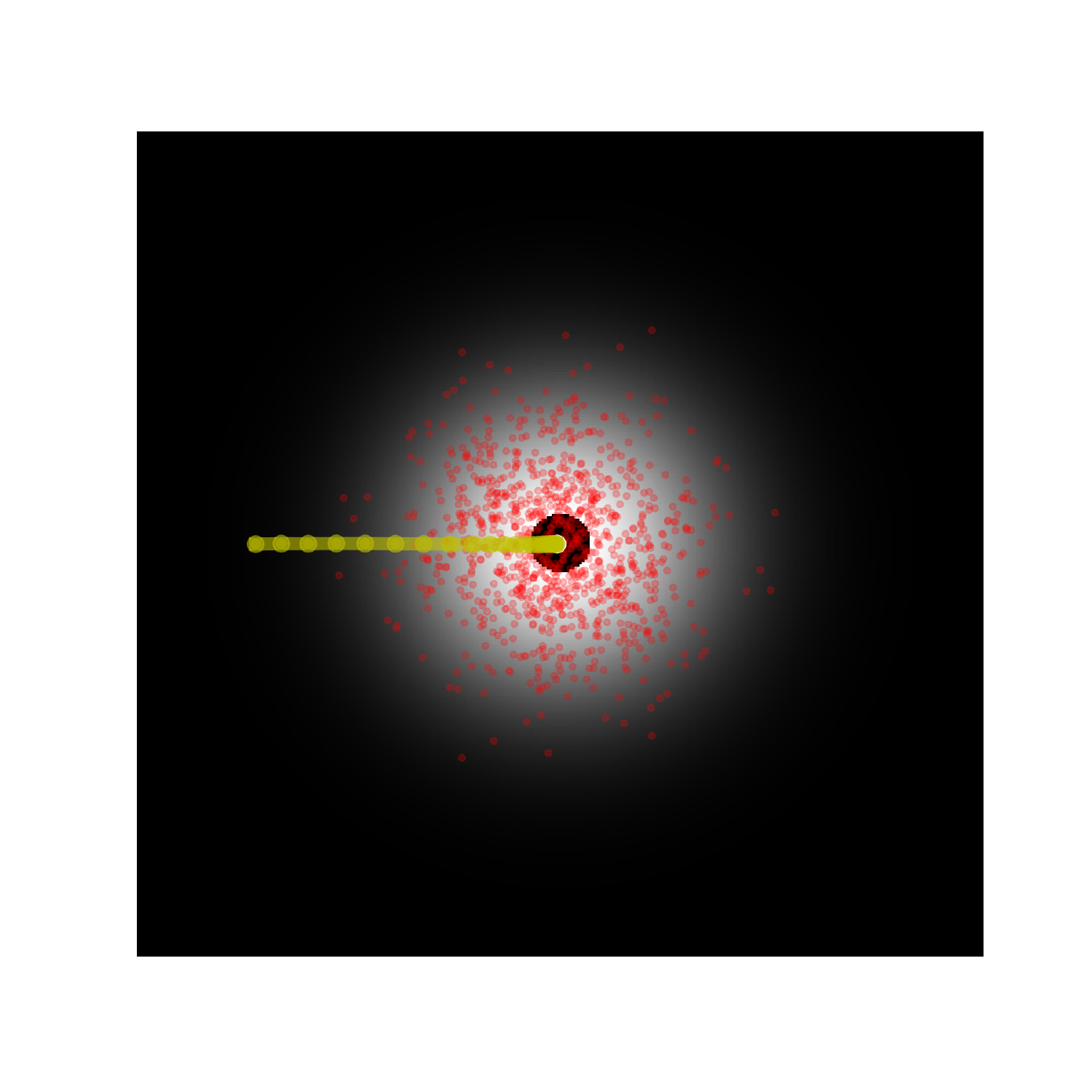}
         \vspace{\cutdistbig}
        \caption{ES with $\sigma=0.16$}
    \end{subfigure}%
  \begin{subfigure}[t]{0.24\textwidth}
        \centering
  \includegraphics[height=\msbig]{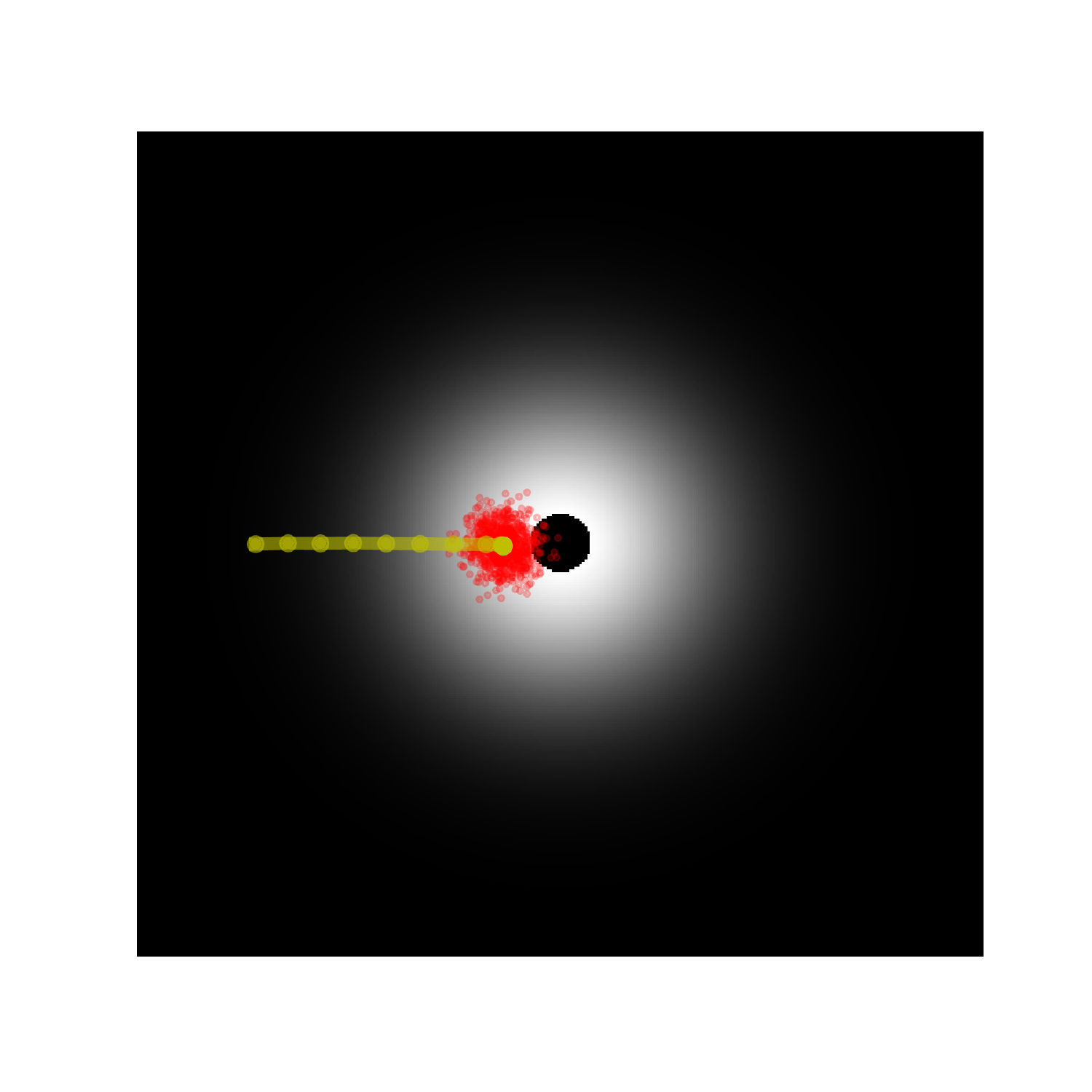}
  \vspace{\cutdistbig}
  \caption{ES with $\sigma=0.04$}
  \end{subfigure} \\
	\vspace{-0.05in}
  \begin{subfigure}[t]{0.24\textwidth}
        \centering
  \includegraphics[height=\msbig]{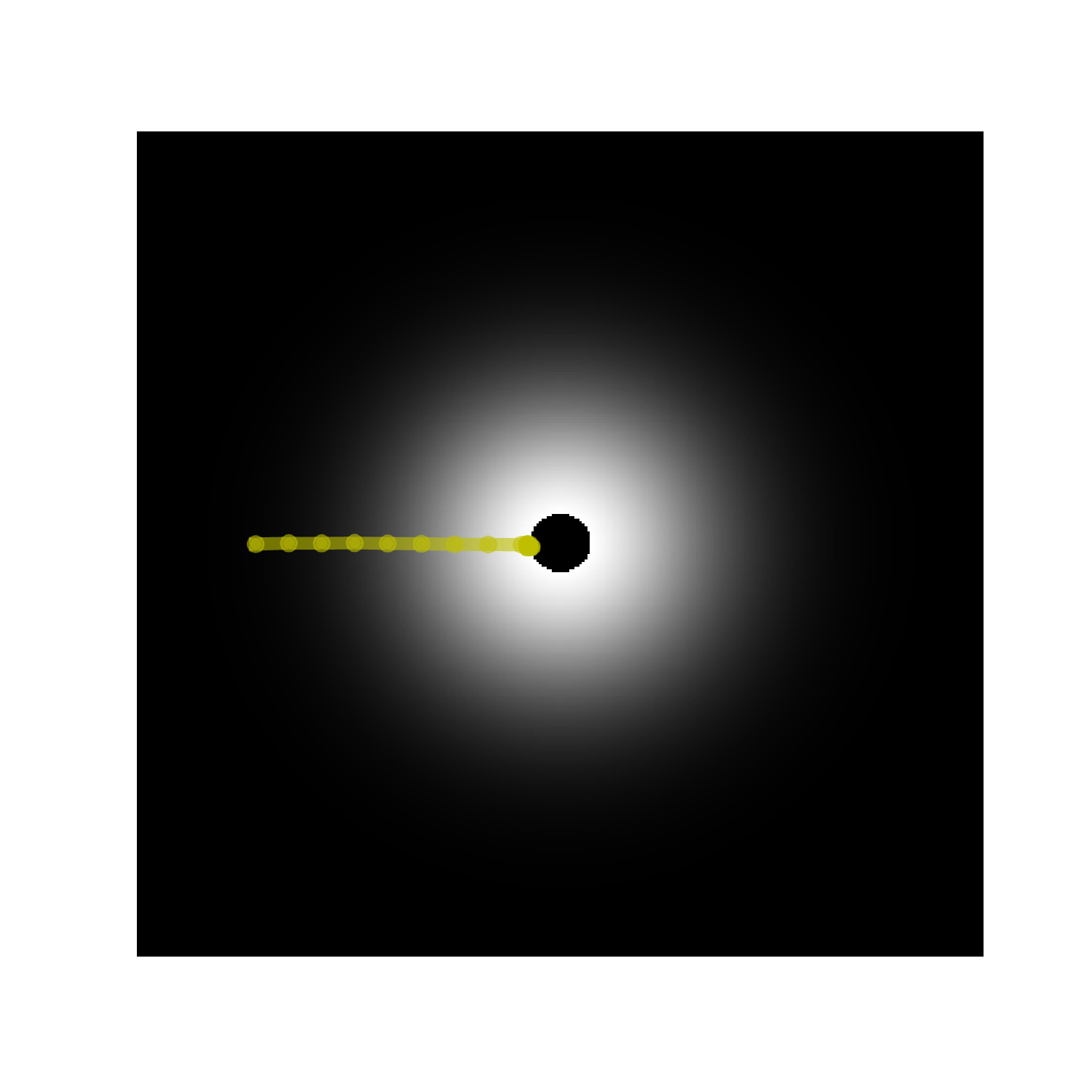}
  \vspace{\cutdistbig}
  \caption{ES with $\sigma=0.002$}
  \end{subfigure}%
  \begin{subfigure}[t]{0.24\textwidth}
        \centering
  \includegraphics[height=\msbig]{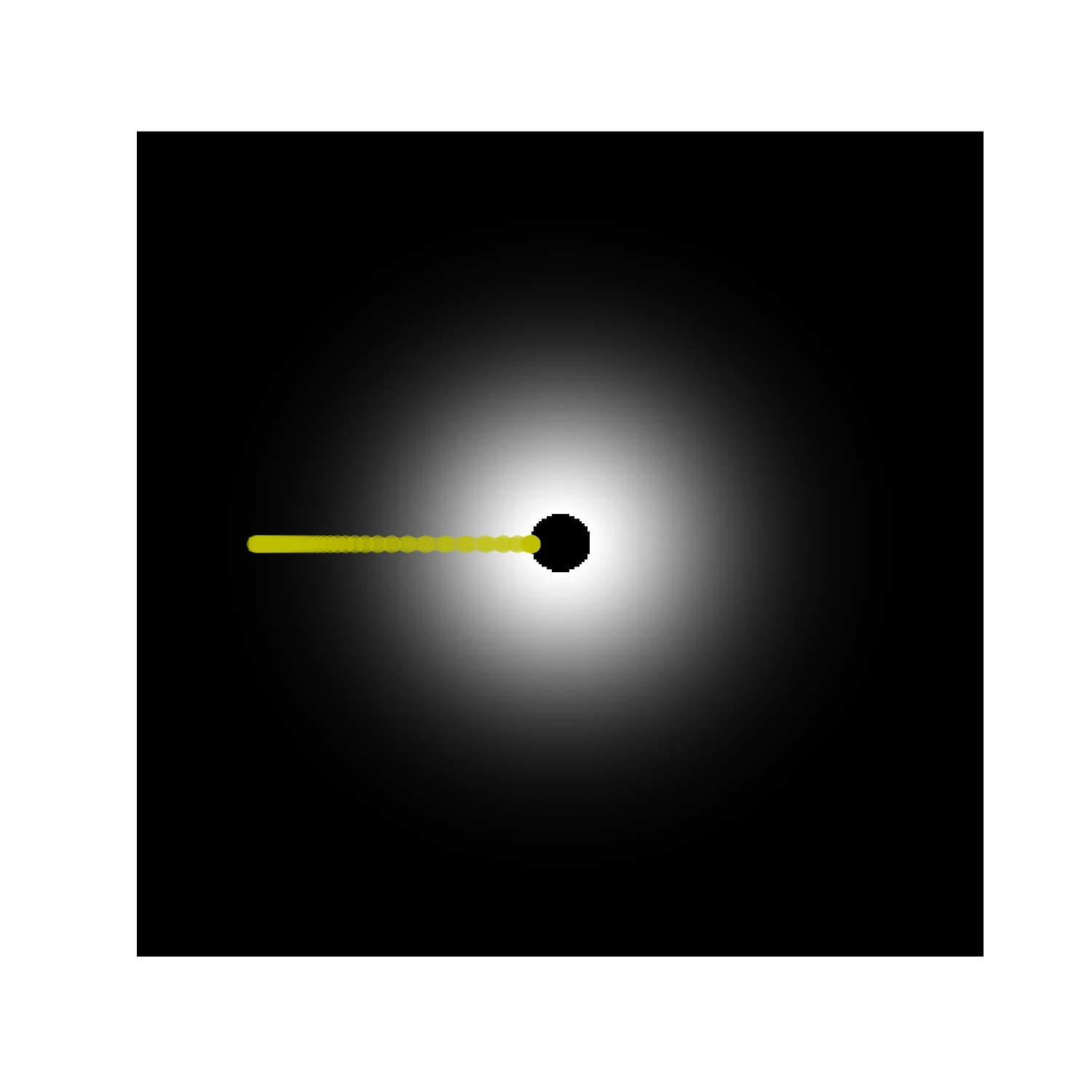}
  \vspace{\cutdistbig}
  \caption{FD with $\epsilon=1e-7$}
  \end{subfigure}
  \vspace{-0.25in}
  \caption{\textbf{Search trajectory comparison in the Donut landscape.} The plots compare representative trajectories of ES with decreasing variance to finite-differences gradient descent. With (a) high variance, ES maximizes expected fitness by moving the distribution's mean into a low-fitness area. With (b,c) decreasing variance, ES is drawn closer to the edge of the low-fitness area, qualitatively converging to the behavior of (d) finite-difference gradient descent.\label{fig:donut}}
\vspace{-0.1in}
%
%
    
\end{figure}

\begin{figure}
  \centering
  \begin{subfigure}[t]{0.49\textwidth}
  \centering
  \includegraphics[height=1.9in]{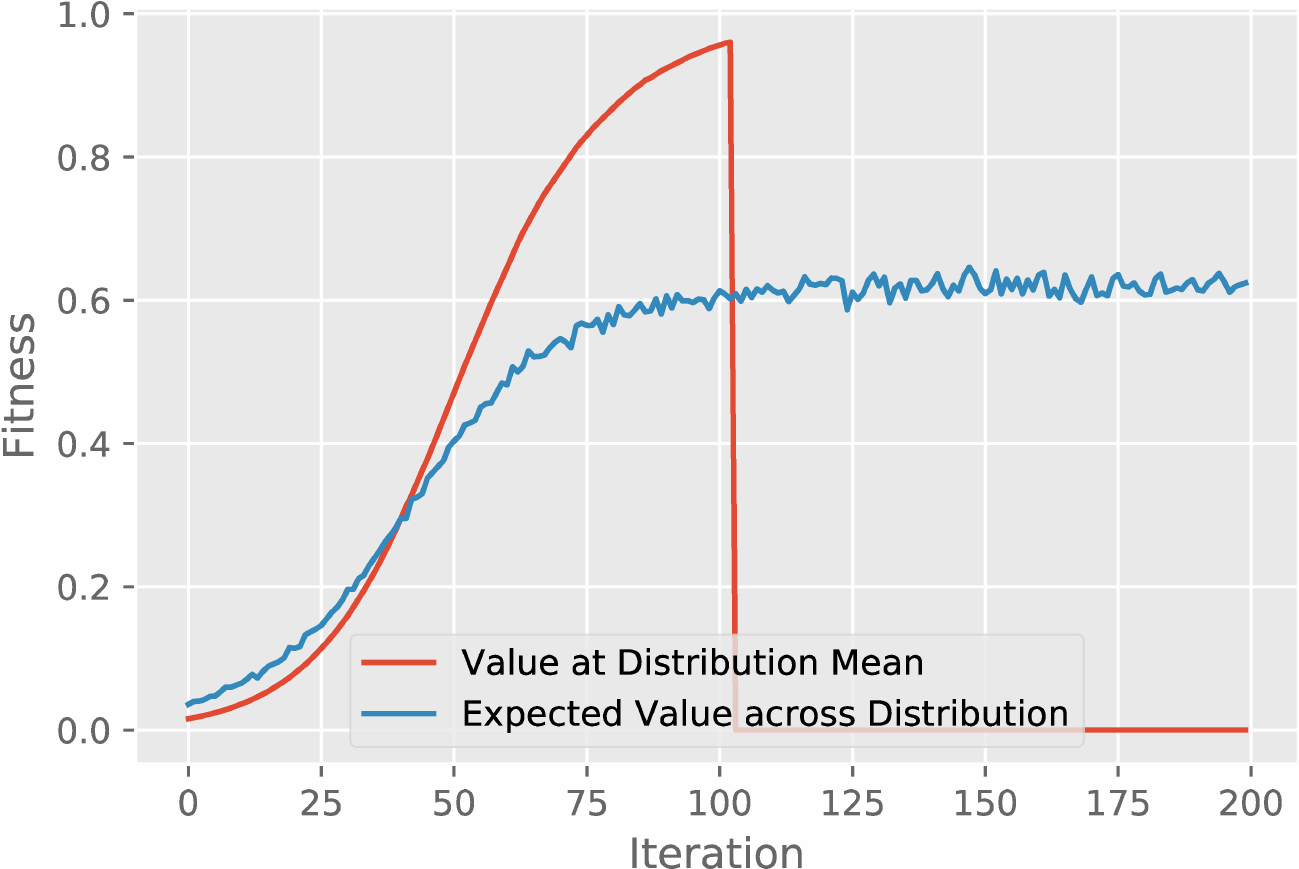}
	    \vspace{-0.05in}
  \caption{Reward of ES on the Donut Landscape}
  \end{subfigure} \\ \vspace{0.1in}
  \begin{subfigure}[t]{0.49\textwidth}
  \centering
  \includegraphics[height=1.9in]{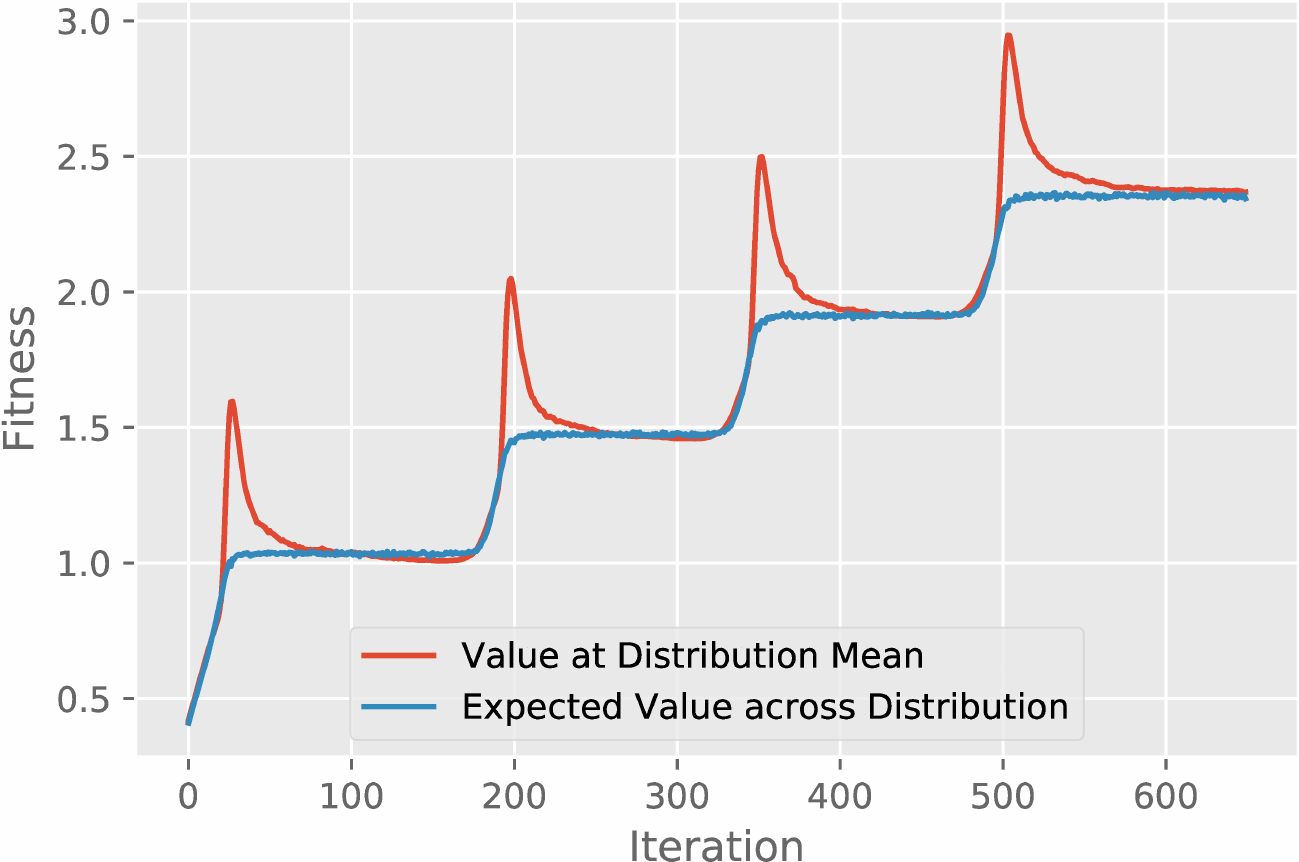}
	    \vspace{-0.05in}
  \caption{Reward of ES on the Fleeting Peaks Landscape}
  \end{subfigure} \\ \vspace{0.1in}
    \begin{subfigure}[t]{0.49\textwidth}
  \centering
  \includegraphics[height=1.9in]{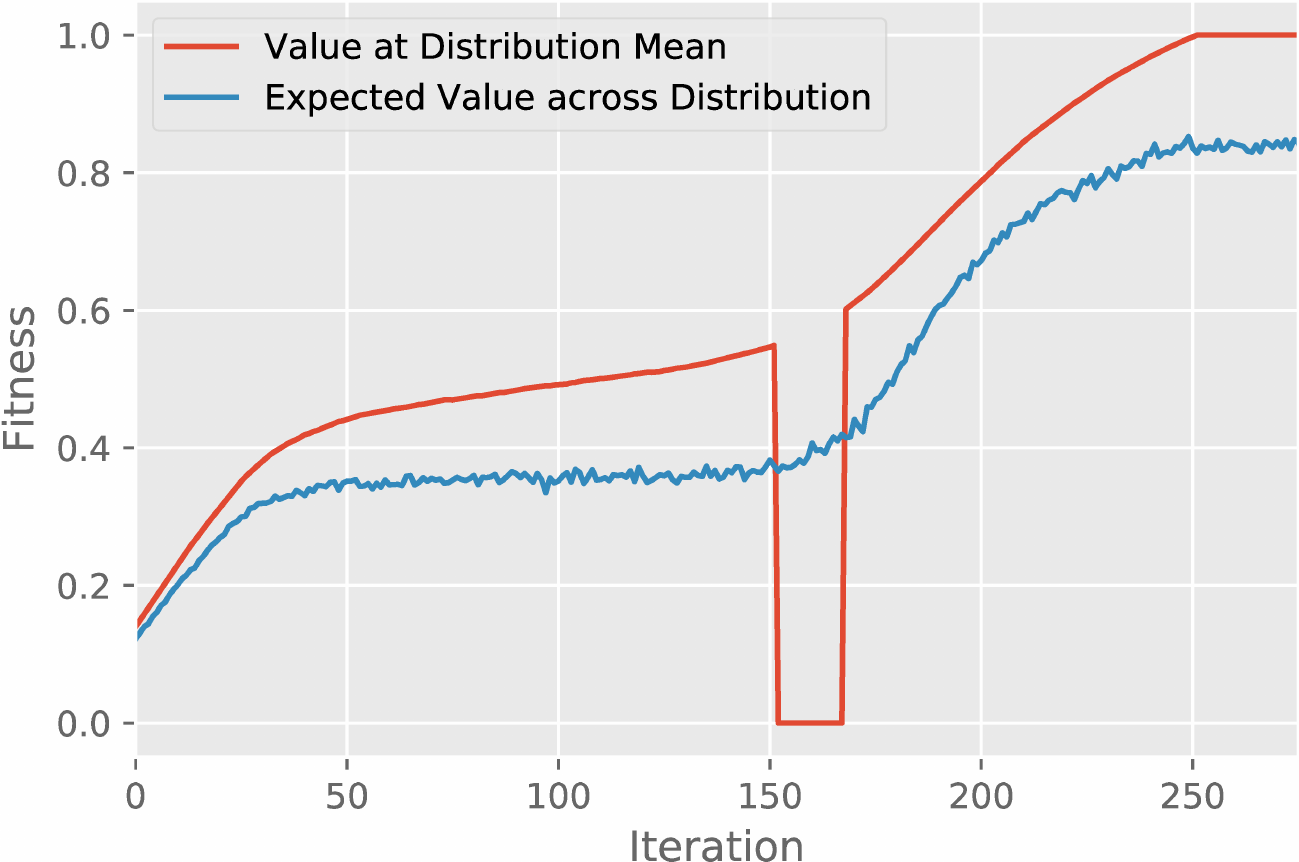}
	    \vspace{-0.05in}
  \caption{Reward of ES on the Gradient Gap Landscape}
  \end{subfigure}
	\vspace{-0.1in}
\caption{\textbf{ES maximizes expected value over the search distribution.} These plots show how the expected value of fitness and the fitness value evaluated at the distribution's mean can diverge in representative runs of ES. This divergence is shown on (a) the Donut landscape with \emph{high} variance ($\sigma=0.16$), (b) the Fleeting Peaks landscape with \emph{medium} variance ($\sigma=0.048$), and  (c) the Gradient Gap landscape with \emph{high} variance ($\sigma=0.18$).
\label{fig:donut_ev}}
	\vspace{-0.1in}
\end{figure}

In the Donut landscape (figure \ref{fig:landscape}a), when the variance of ES's Gaussian is high enough (i.e.\ $\sigma$ of the search distribution is set to $0.16$, shown in figure \ref{fig:donut}a), ES maximizes distributional reward by centering the mean of its domain parameter distribution
at the middle of the donut where fitness is \emph{lowest}; figure \ref{fig:donut_ev}a further illuminates this
divergence. When ES's variance is smaller ($\sigma = 0.04$), ES instead positions itself such that the tail of its distribution avoids the donut hole (figure \ref{fig:donut}b). Finally, when ES's variance becomes tiny ($\sigma = 0.002$), the distribution becomes tightly distributed along the edge of the donut-hole (figure \ref{fig:donut}c). This final ES behavior is qualitatively similar to following a FD approximation of the domain parameter performance gradient (figure \ref{fig:donut}d). 

In the Narrowing Path landscape (figure \ref{fig:landscape}b), when ES is applied with high variance ($\sigma=0.12$) it is unable to progress far along the narrowing path to higher fitness (figure \ref{fig:narrowing_path}a), because expected value is highest when a significant portion of the distribution remains on the path. As variance declines (figures \ref{fig:narrowing_path}b and \ref{fig:narrowing_path}c), ES proceeds further along the path. FD gradient descent is able to easily traverse the entire path (figure \ref{fig:narrowing_path}d).

\def \Amsbig {1.7in}

\begin{figure}[h]
  \centering
  \begin{subfigure}[t]{0.24\textwidth}
  		\centering
         \includegraphics[height=\Amsbig]{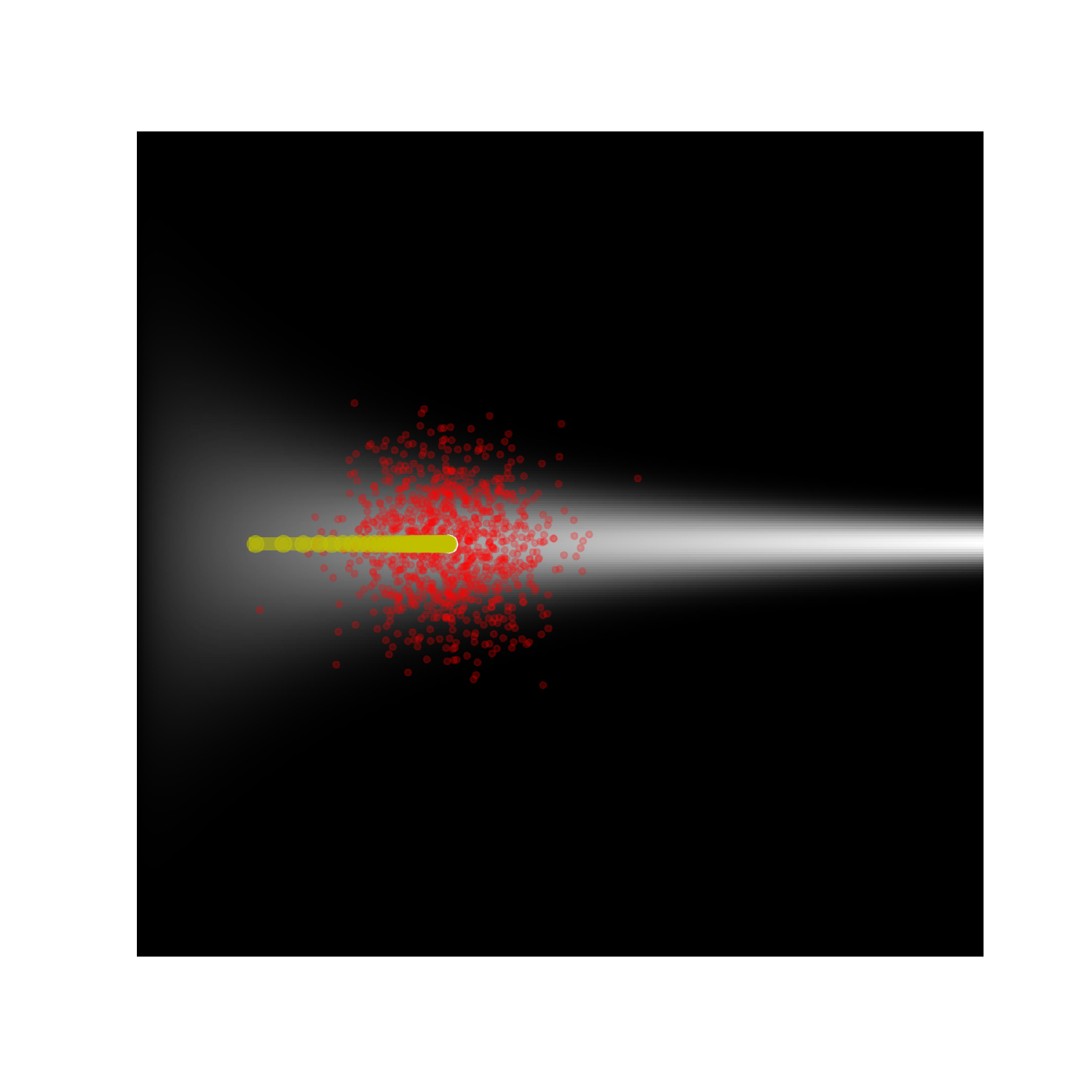}
         \vspace{\cutdistbig}
        \caption{ES with $\sigma=0.12$}
    \end{subfigure}%
  \begin{subfigure}[t]{0.24\textwidth}
        \centering
  \includegraphics[height=\Amsbig]{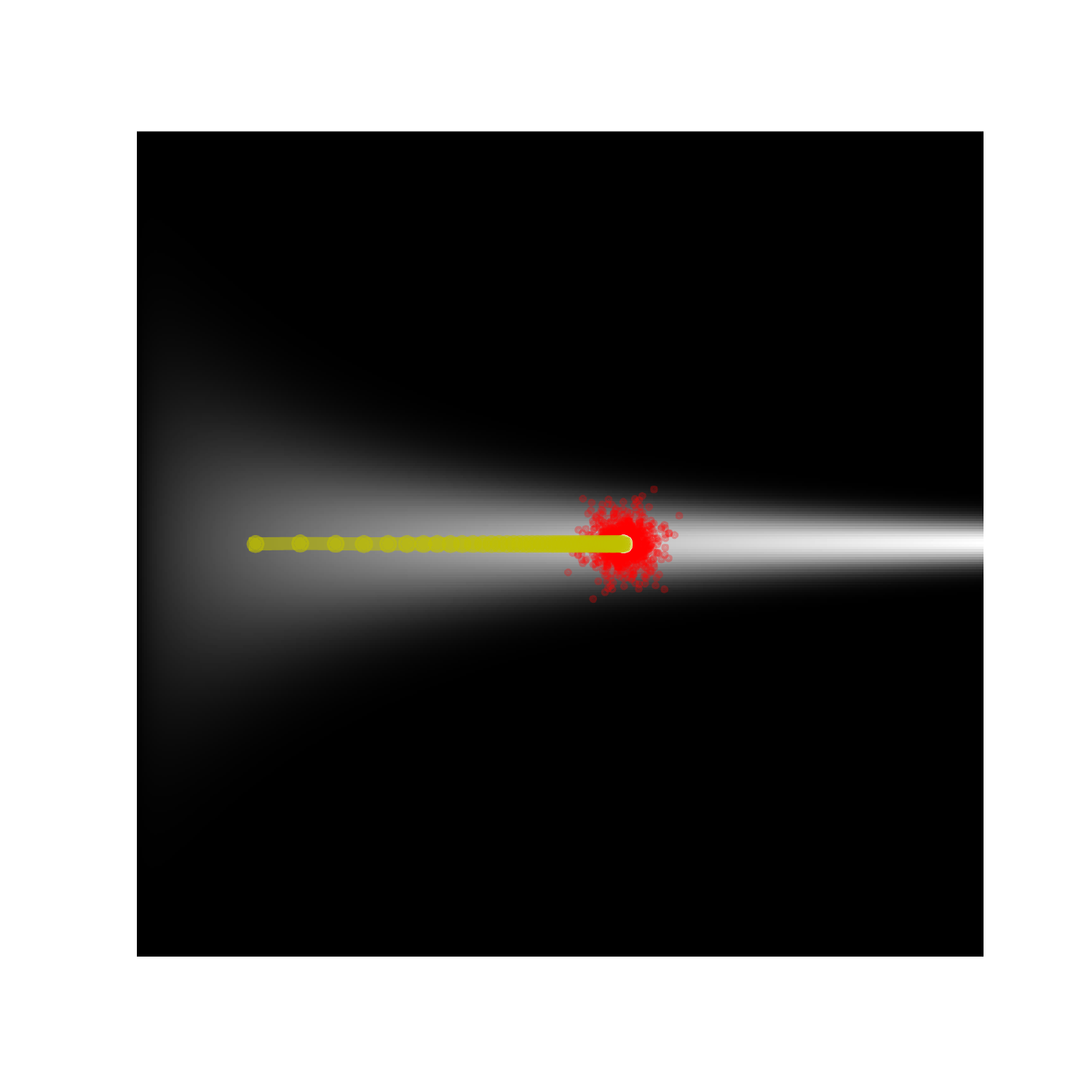}
           \vspace{\cutdistbig}
  \caption{ES with $\sigma=0.04$}
  \end{subfigure} \\
	\vspace{-0.05in}
 \begin{subfigure}[t]{0.24\textwidth}
        \centering
  \includegraphics[height=\Amsbig]{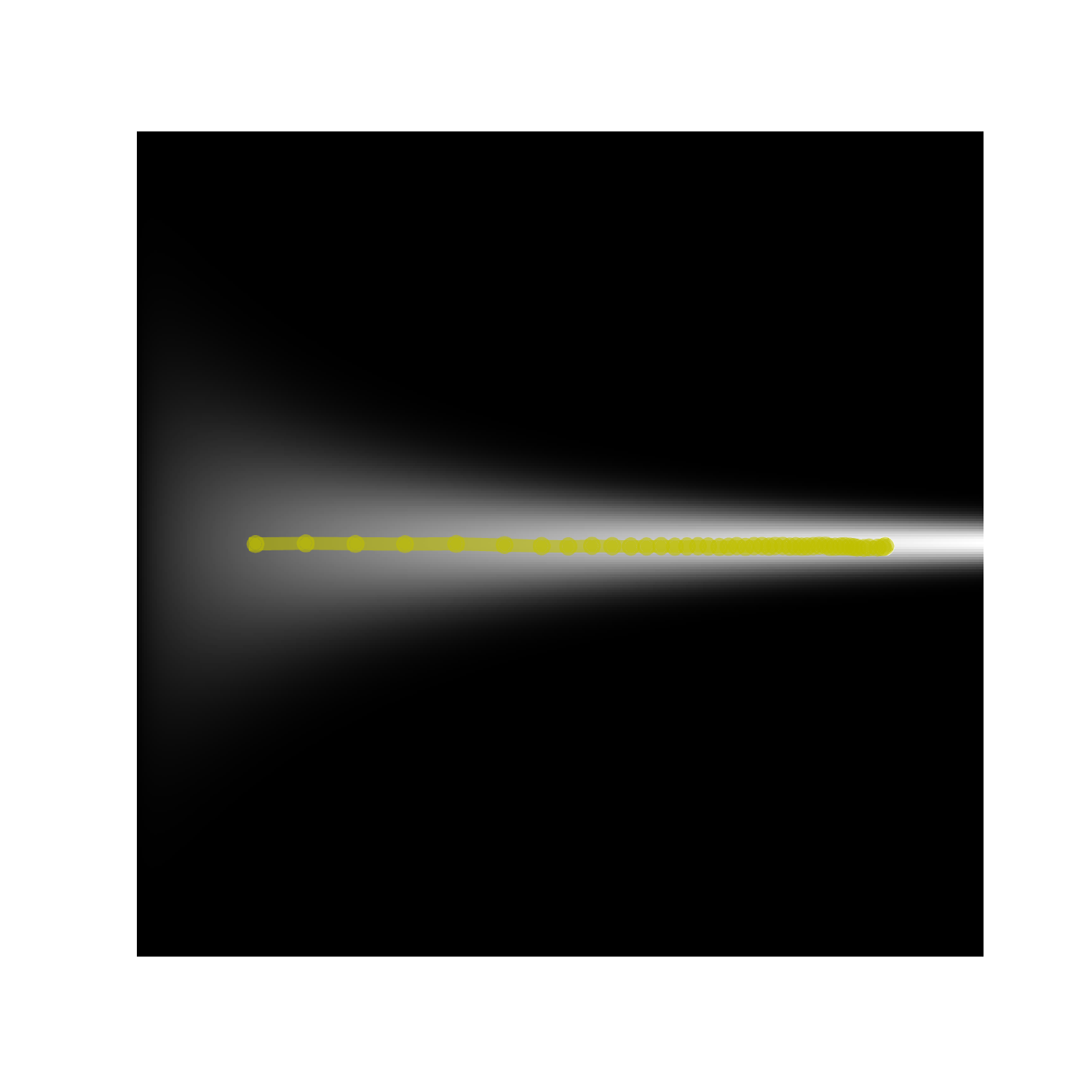}
           \vspace{\cutdistbig}
  \caption{ES with $\sigma=0.0005$}
  \end{subfigure}%
   \begin{subfigure}[t]{0.24\textwidth}
        \centering
  \includegraphics[height=\Amsbig]{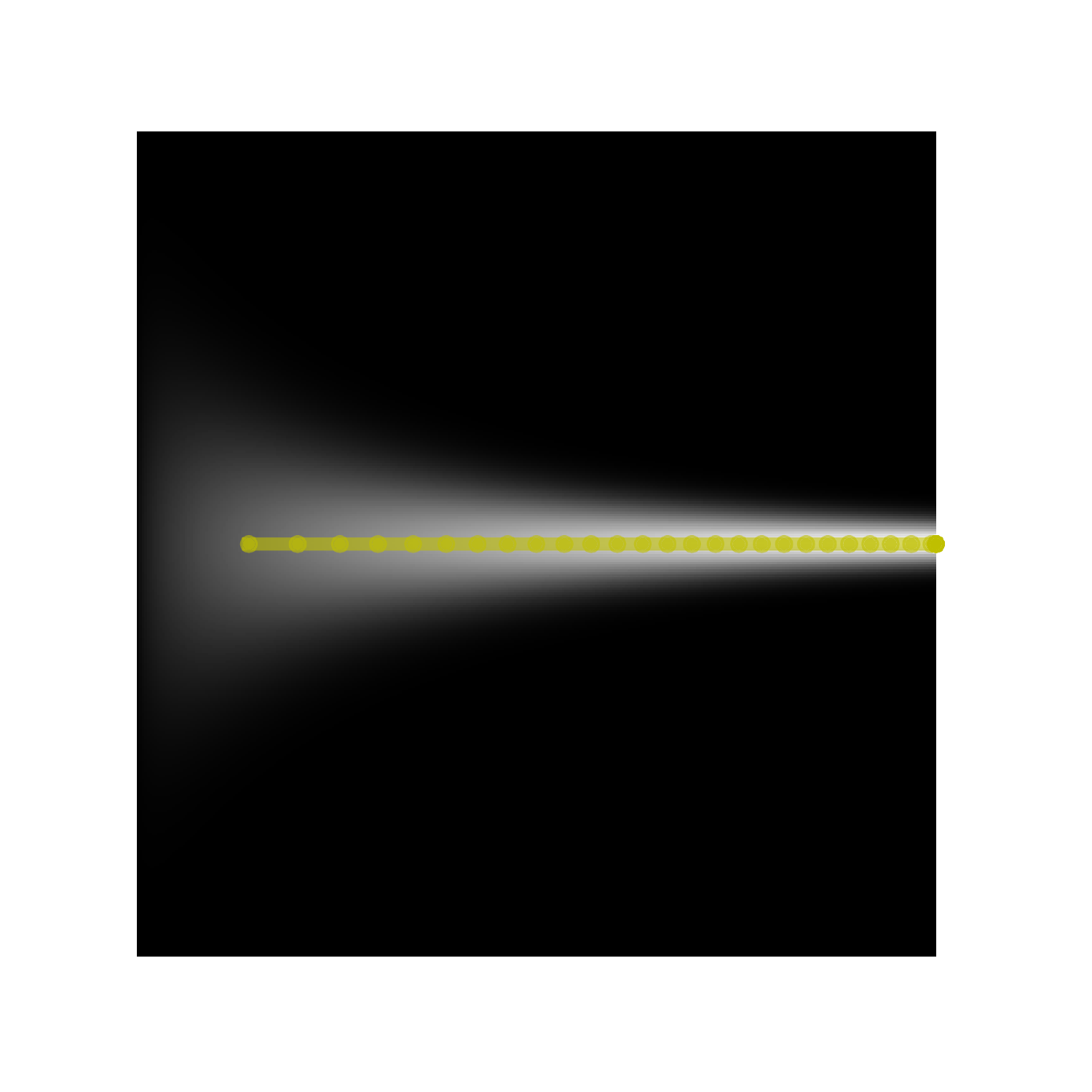}
           \vspace{\cutdistbig}
  \caption{FD with $\epsilon=1e-7$}
  \end{subfigure}
  \vspace{-0.2in} 
	\caption{\textbf{Search trajectory comparison in the Narrowing Path landscape.} With (a) high variance, ES cannot proceed as the path narrows because its distribution increasingly falls outside the path. Importantly, if ES is being used to ultimately discover a single high-value policy, as is often the case \citep{salimans:es}, this method will not discover the superior solutions further down the path. With (b,c) decreasing variance, ES is able to traverse further along the narrowing path. (d) FD gradient descent traverses the entire path.\label{fig:narrowing_path}}
	\vspace{-0.05in}
\end{figure}

\begin{figure}[h]
  \centering
  \begin{subfigure}[t]{0.24\textwidth}
  		\centering
         \includegraphics[height=\msbig]{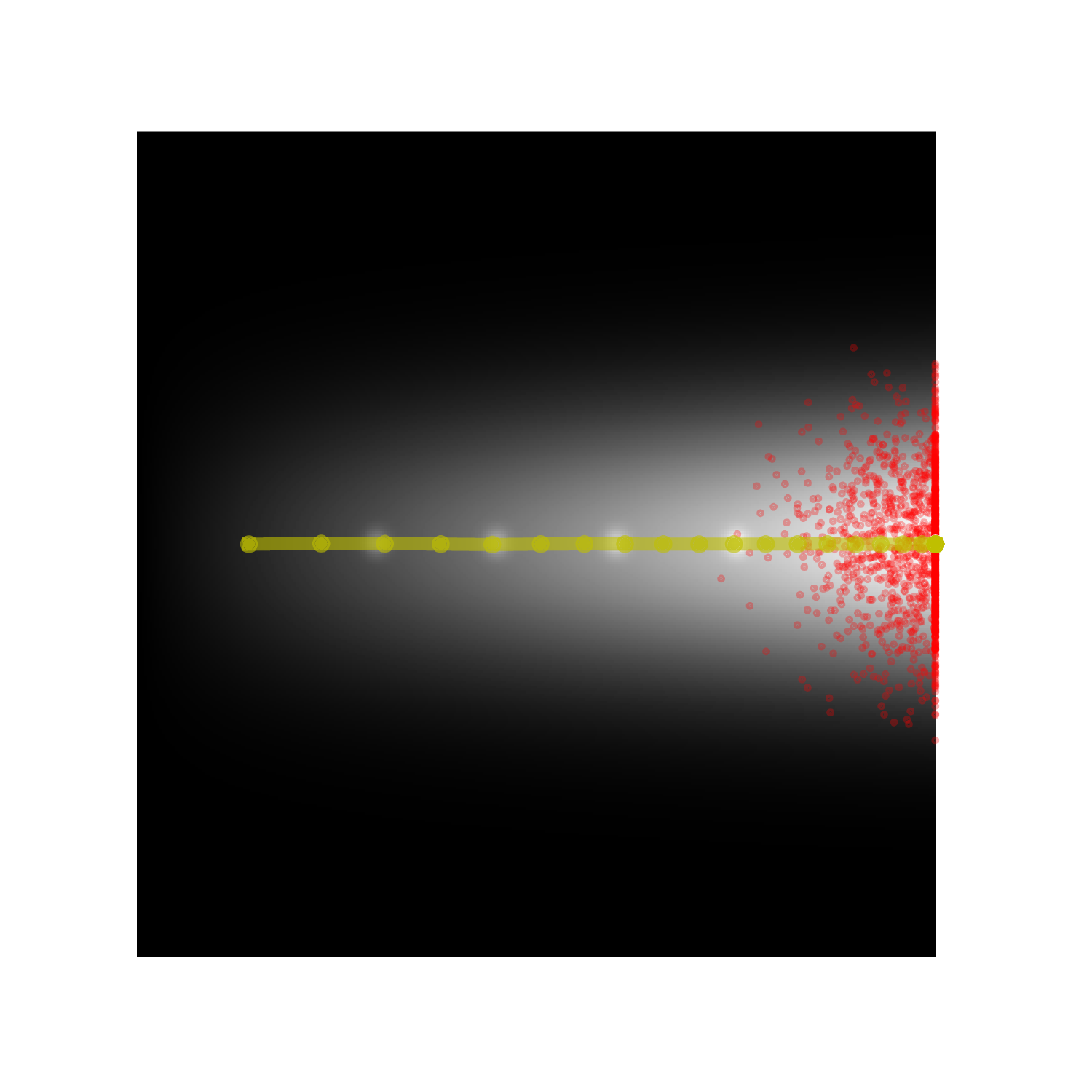}
           \vspace{\cutdistbig}
        \caption{ES with $\sigma=0.16$}
    \end{subfigure}%
  \begin{subfigure}[t]{0.24\textwidth}
        \centering
  \includegraphics[height=\msbig]{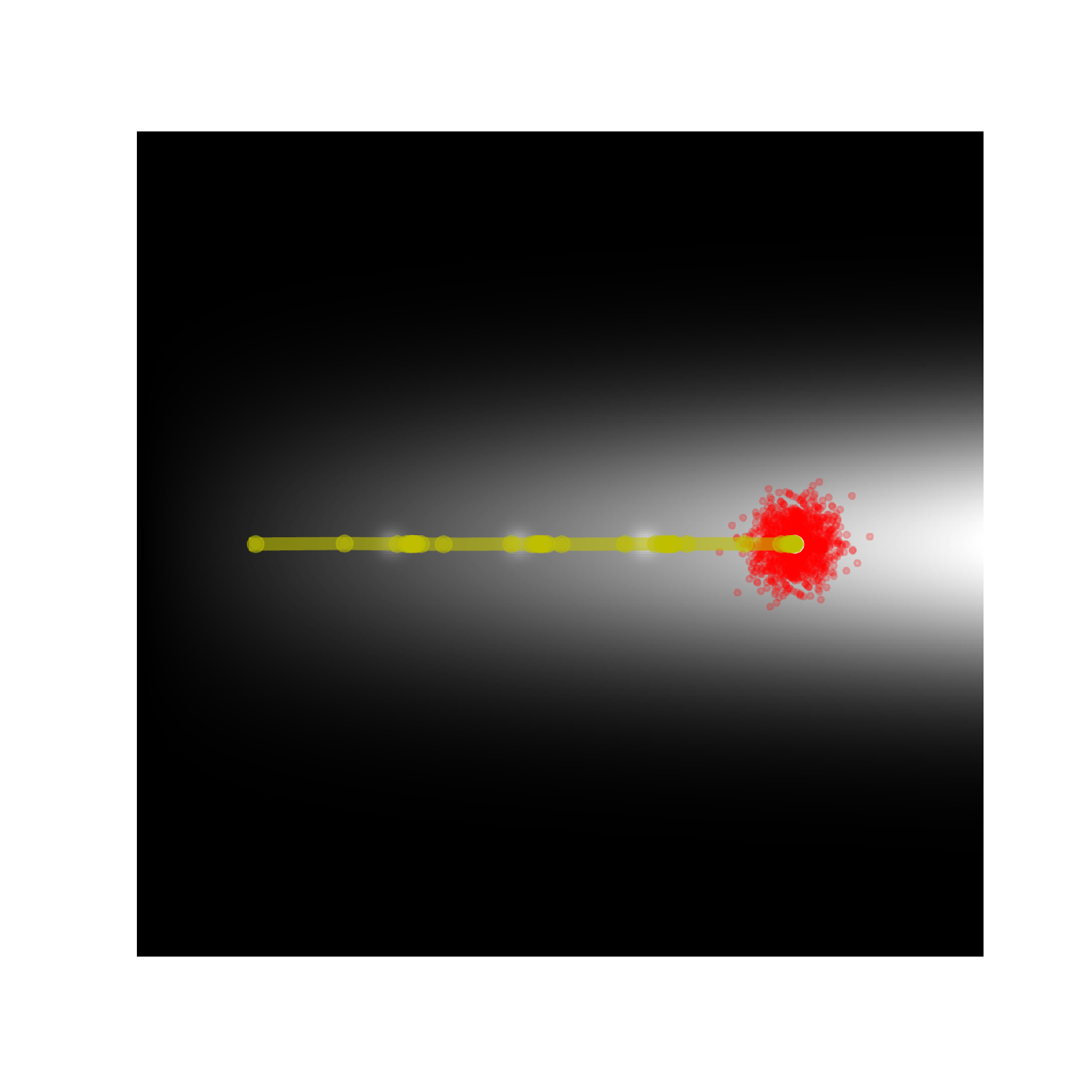}
    \vspace{\cutdistbig}
  \caption{ES with $\sigma=0.048$}
  \end{subfigure} \\
	\vspace{-0.05in}
 \begin{subfigure}[t]{0.24\textwidth}
        \centering
  \includegraphics[height=\msbig]{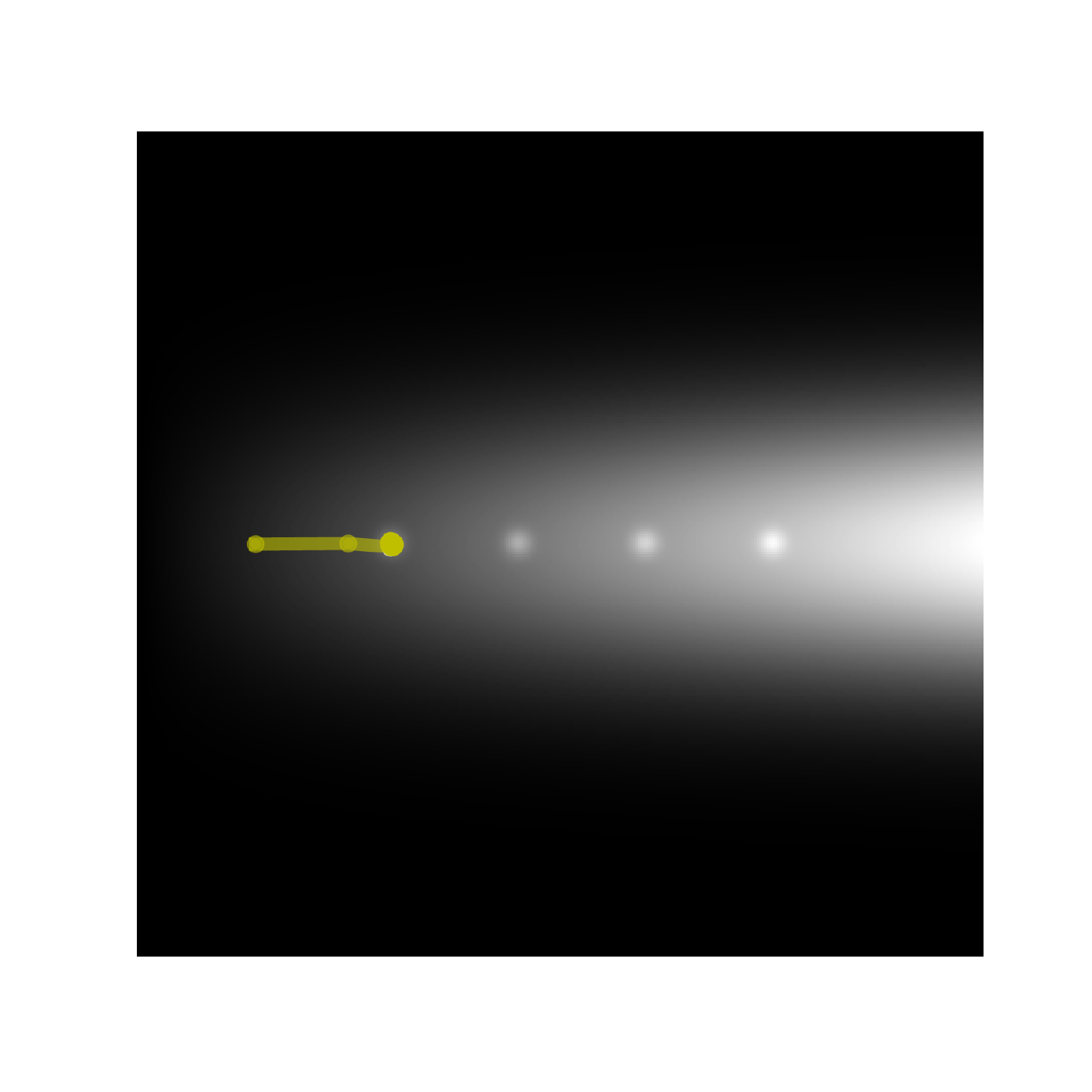}
    \vspace{\cutdistbig}
  \caption{ES with $\sigma=0.002$}
  \end{subfigure}%
   \begin{subfigure}[t]{0.24\textwidth}
        \centering
  \includegraphics[height=\msbig]{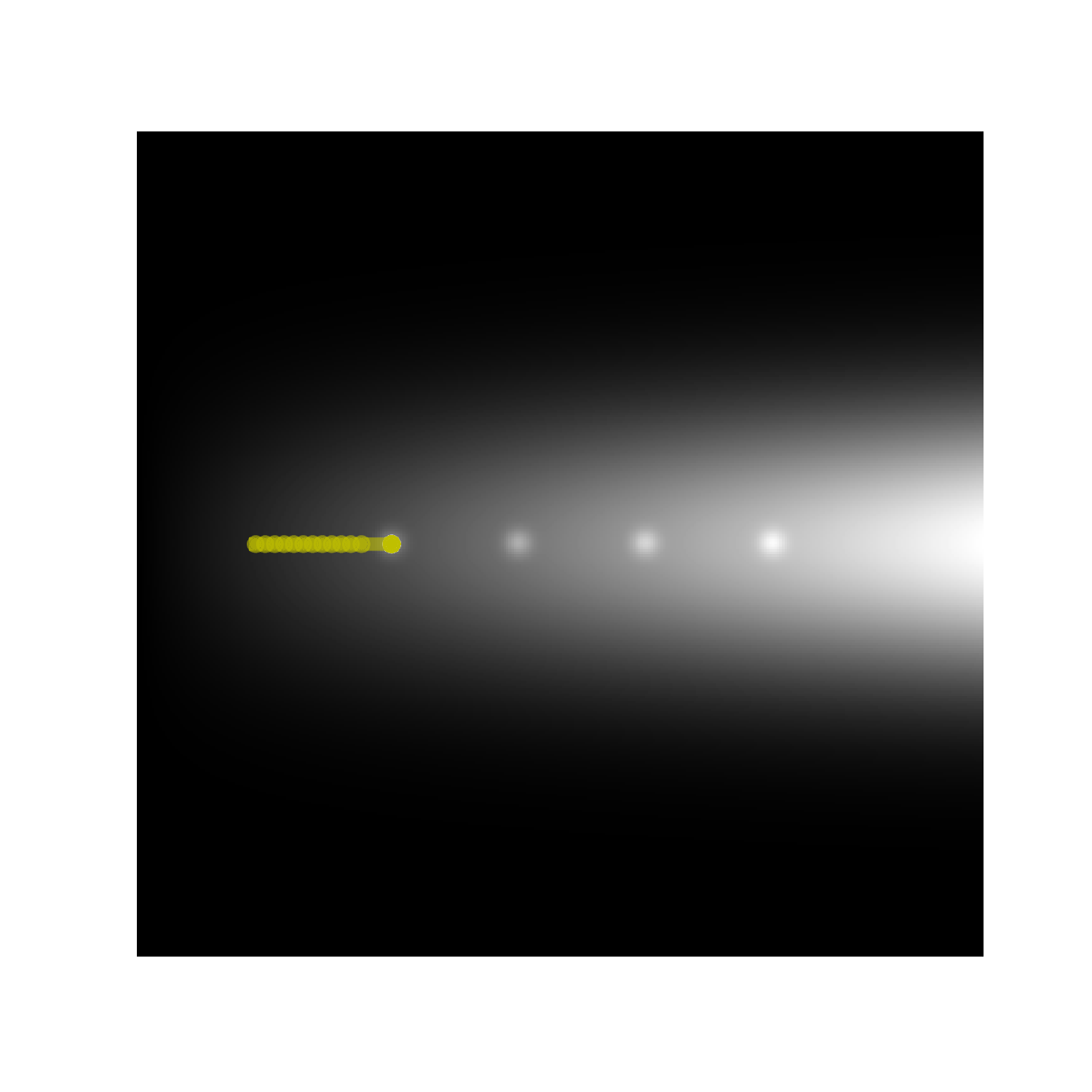}
    \vspace{\cutdistbig}
  \caption{FD with $\epsilon=1e-7$}
  \end{subfigure}
  \vspace{-0.3in} 
  \caption{\textbf{Search trajectory comparison in the Fleeting Peaks landscape.} With (a) high variance, ES ignores the local optima because of their relatively small contribution to expected fitness. With (b) medium variance, ES hops between local optima, and with (c) low variance, ES converges to a local optimum, as does (d) FD gradient descent. \label{fig:fleeting_peaks}}
	\vspace{-0.15in}
\end{figure}

In the Fleeting Peaks landscape (figure \ref{fig:landscape}c), when high-variance ES is applied ($\sigma=0.16$) the search distribution has sufficient spread to ignore the local optima and proceeds to the maximal-fitness area (figure \ref{fig:fleeting_peaks}a). With medium variance ($\sigma=0.048$; figure \ref{fig:fleeting_peaks}b), ES gravitates to each local optima before leaping to the next one, ultimately becoming stuck on the last local optimum (see figure \ref{fig:donut_ev}b). With low variance ($\sigma=0.002$; figure \ref{fig:fleeting_peaks}c), ES latches onto the first local optimum and remains stuck there indefinitely; FD gradient descent becomes similarly stuck (figure \ref{fig:fleeting_peaks}d).

\begin{figure*}[h]
  \centering
  \begin{subfigure}[t]{0.29\textwidth}
  		\centering
	  \includegraphics[height=\msz]{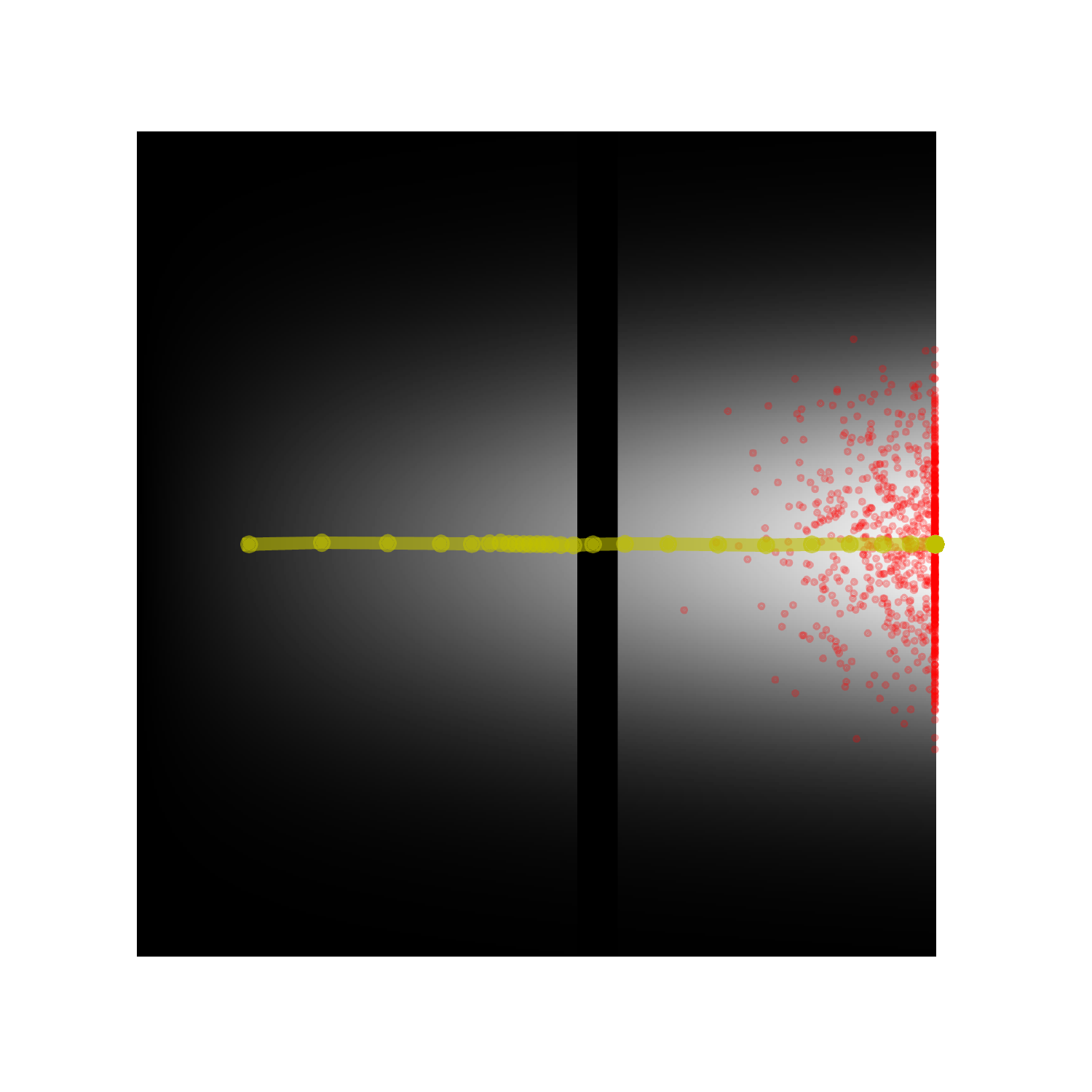}
         \vspace{-0.25in}
        \caption{ES with $\sigma=0.18$}
    \end{subfigure}%
  \begin{subfigure}[t]{0.29\textwidth}
        \centering
  \includegraphics[height=\msz]{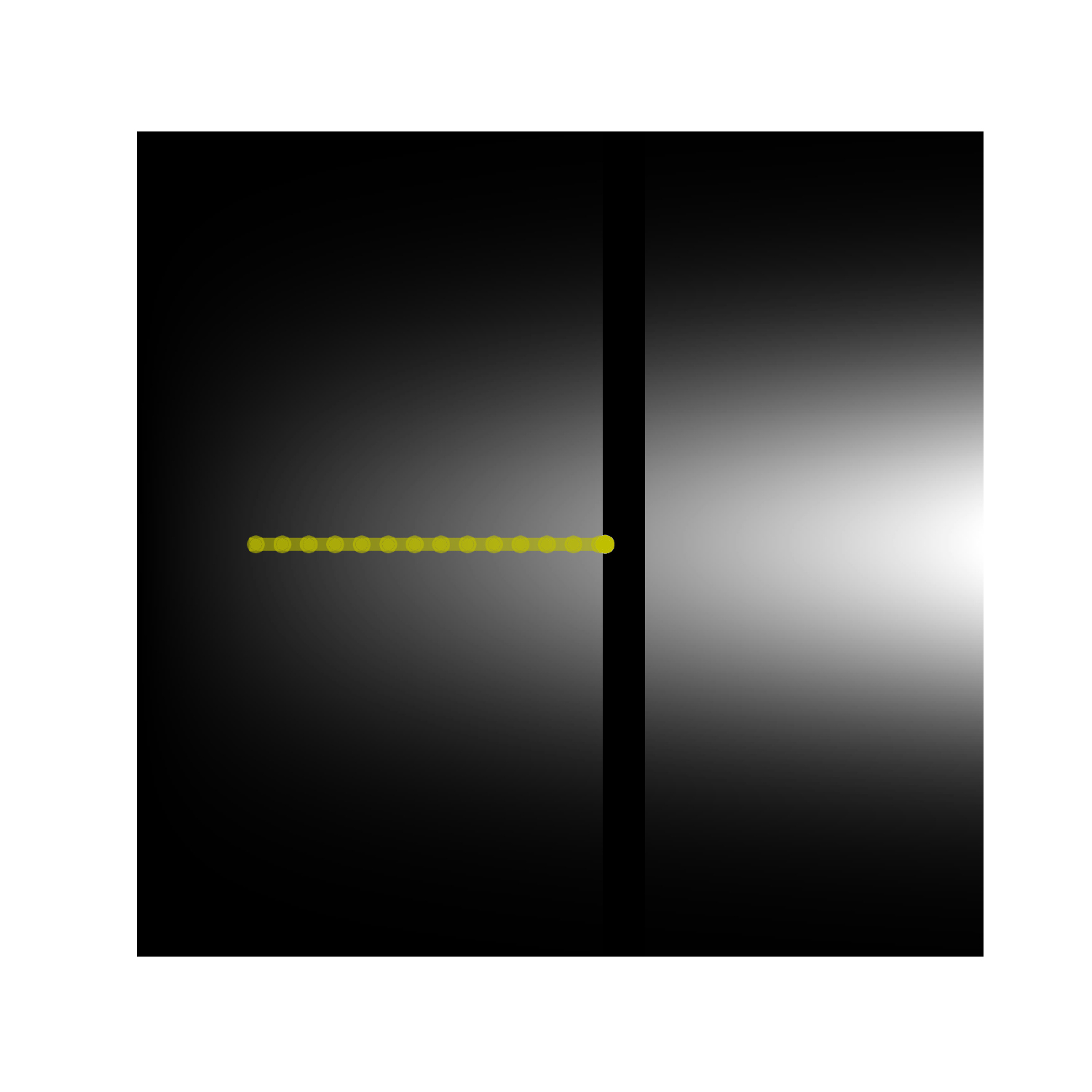}
  \vspace{-0.25in}
  \caption{Finite Differences}
  \end{subfigure}
  \begin{subfigure}[t]{0.29\textwidth}
        \centering
  \includegraphics[height=\msz]{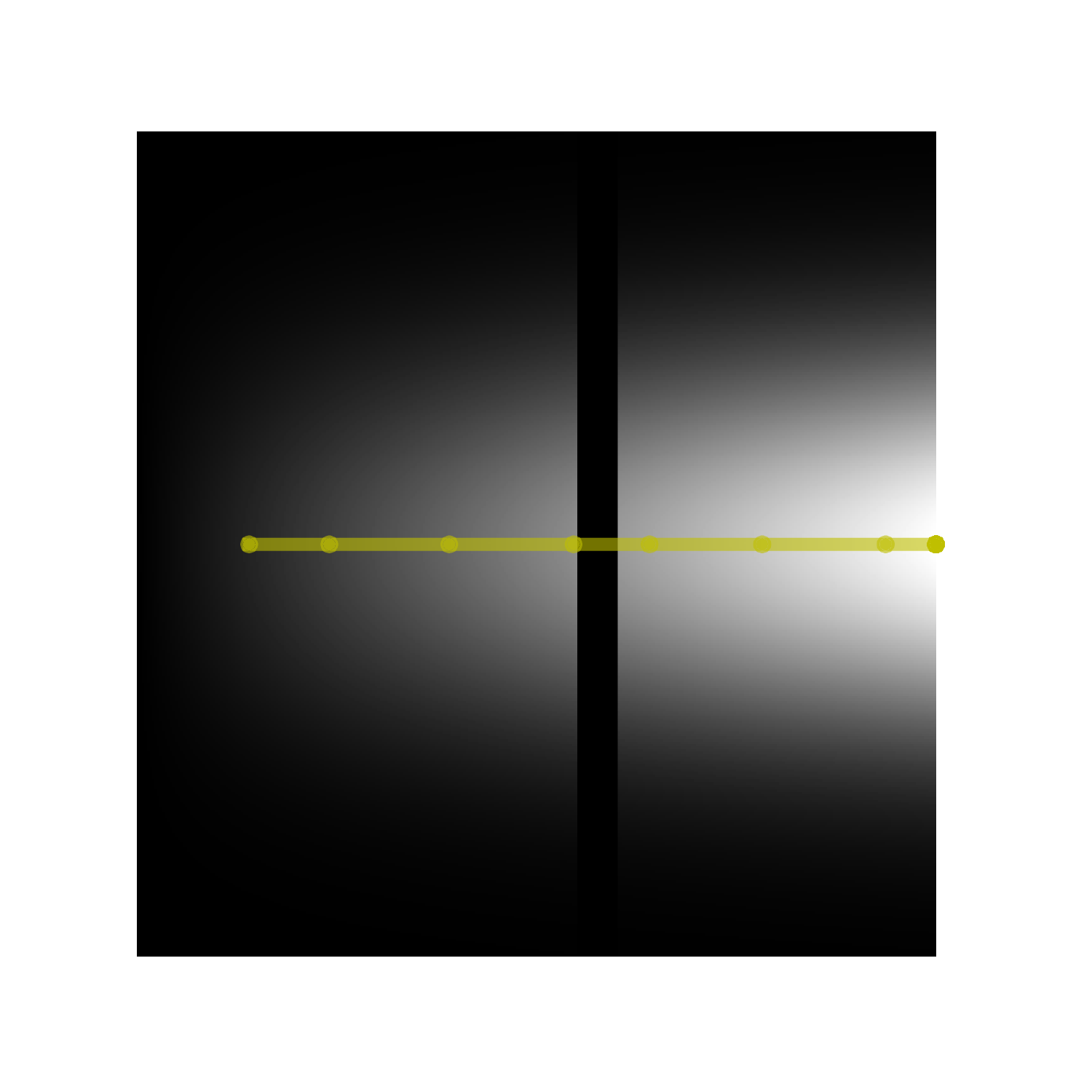}
 \vspace{-0.25in}
  \caption{Finite Differences + Momentum}
  \end{subfigure} \\ \vspace{-0.05in} 
 \begin{subfigure}[t]{0.29\textwidth}
        \centering
	 \includegraphics[height=\msz]{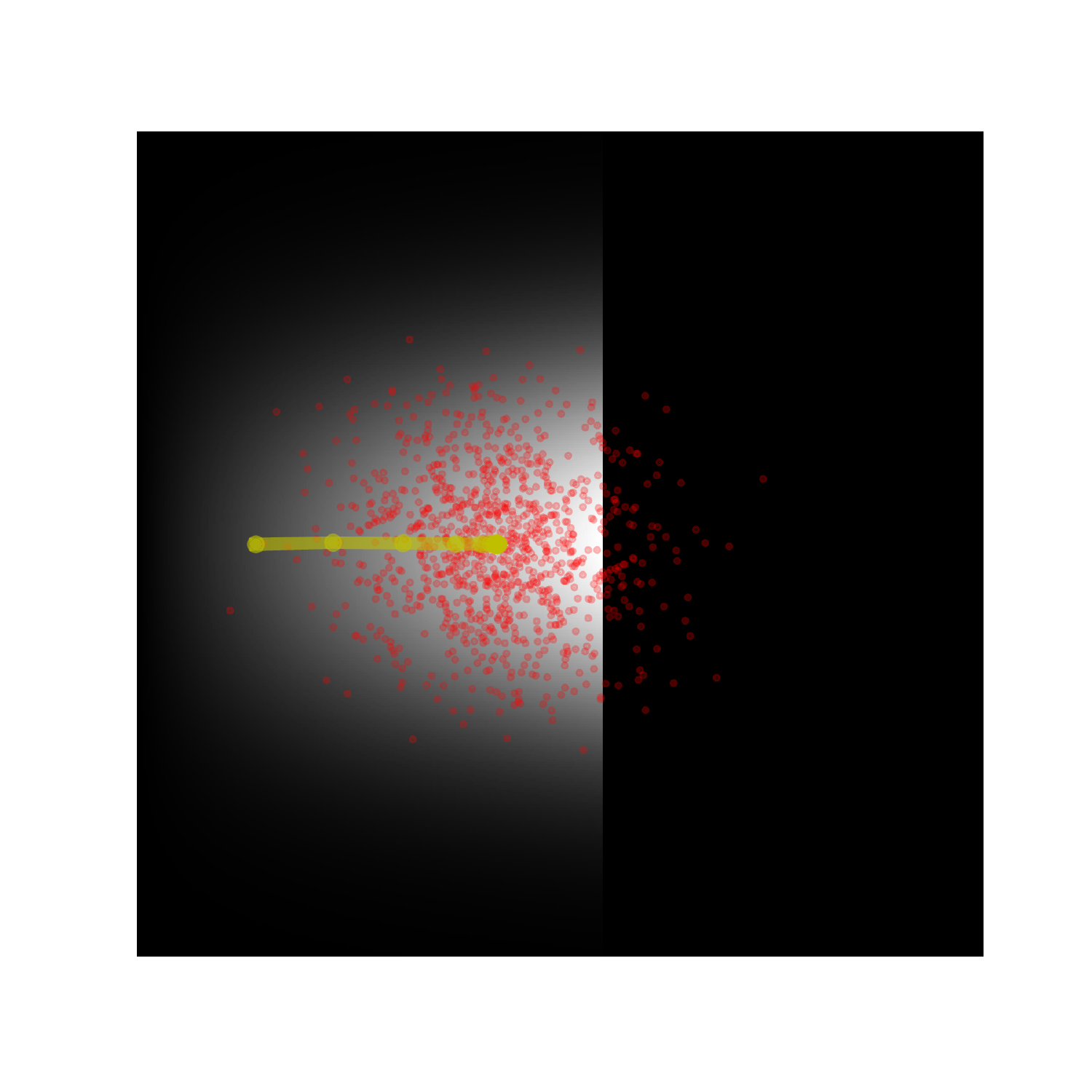}
 \vspace{-0.25in}
  \caption{ES with $\sigma=0.18$}
  \end{subfigure}
   \begin{subfigure}[t]{0.29\textwidth}
        \centering
  \includegraphics[height=\msz]{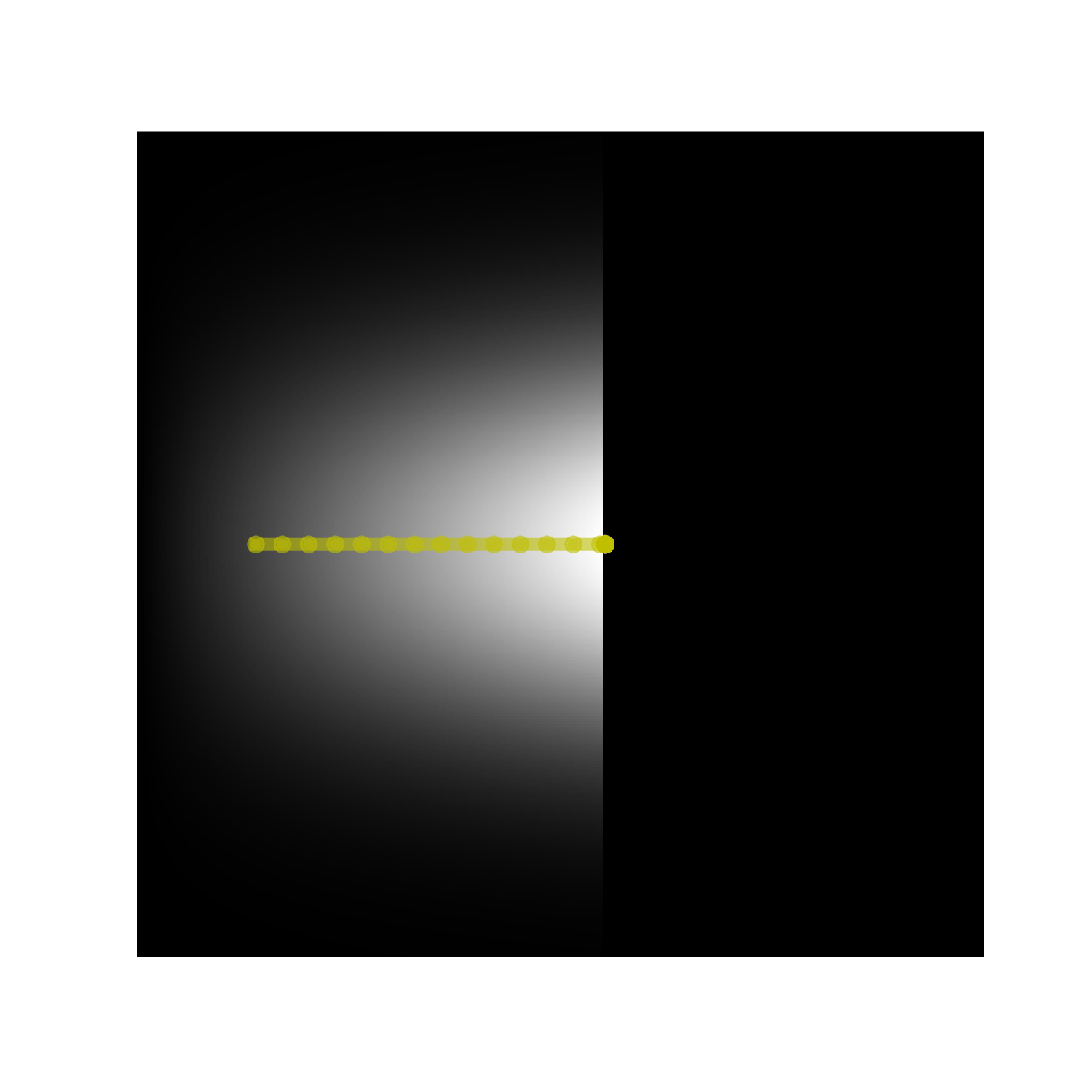}
 \vspace{-0.25in}
  \caption{Finite Differences}
 \end{subfigure}
    \begin{subfigure}[t]{0.29\textwidth}
        \centering
  \includegraphics[height=\msz]{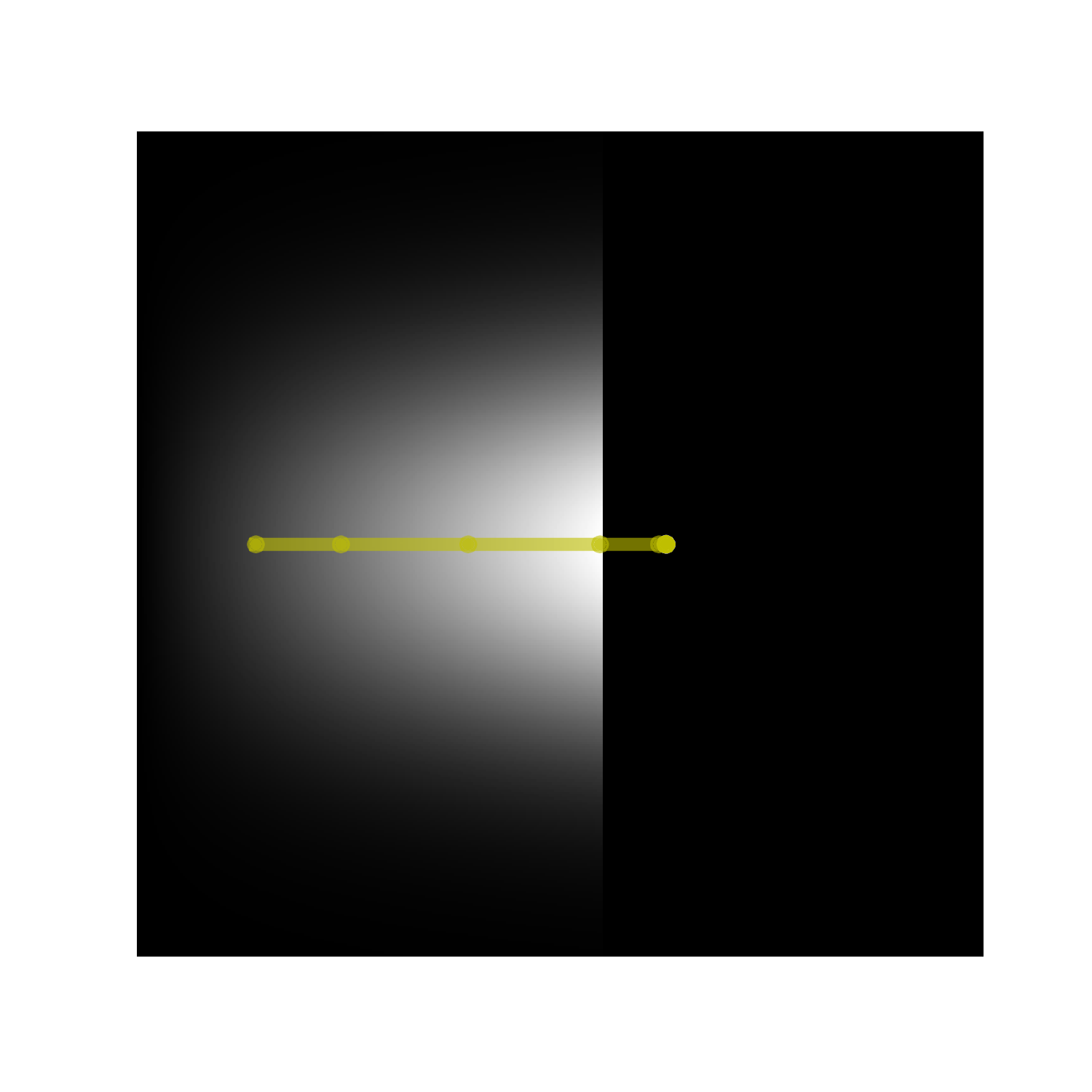}
  \vspace{-0.25in}
  \caption{Finite Differences + Momentum}
  \end{subfigure}
	\vspace{-0.1in}
  \caption{\textbf{Search trajectory comparison in the Gradient Gap and Gradient Cliff landscapes.} With (a) high variance, ES can bypass the gradient-free gap because its distribution can span the gap; with lower-variance ES or (b) FD, search cannot cross the gap. When (c) FD is augmented with momentum, it too can jump across the gap. In the control Gradient Cliff landscape, (d) ES with high variance remains rooted in the high-fitness area, and the performance of (e) FD is unchanged from the Gradient Gap landscape. When (f) FD is combined with momentum, it jumps into the fitness chasm. The conclusion is that only high-variance ES exploits distributional information to make an informed choice about whether or not to move across the gap.
  \label{fig:gradient_gap}}
\end{figure*}

Finally, in the Gradient Gap landscape (figure \ref{fig:landscape}d), ES with high variance ($\sigma=0.18$) can traverse a zero-fitness non-diffentiable gap in the landscape (figure \ref{fig:gradient_gap}a), demonstrating ES's ability to ``look ahead'' in parameter space to  cross fitness valleys between local optima (see also figure \ref{fig:donut_ev}c). Lower variance ES (not shown) and FD cannot cross the gap (figure \ref{fig:gradient_gap}b). Highlighting that ES is informed by samples at the tail of the search distribution and is not blindly pushing forward, ES with high variance in the Gradient Cliff landscape (figure \ref{fig:gradient_gap}d) does not leap into the cliff, and lower variance ES (not shown) and finite differences (figure \ref{fig:gradient_gap}e) behave no different then they do in the Gradient Gap landscape. 

To explore whether more sophisticated update rules could enable FD to behave more similarly to high-variance ES, FD with momentum was additionally applied in both the Gradient Gap and Gradient Cliff landscapes. Momentum is a heuristic often combined with SGD, motivated by the insight that local optima in rugged landscapes can sometimes by avoided by accumulating \emph{momentum} along previously beneficial directions. The question in this experiment is whether such momentum might help FD to cross the zero-fitness area of the Gradient Gap landscape. Indeed, FD with sufficient momentum ($\gtrapprox$0.8) can cross the Gradient Gap (figure \ref{fig:gradient_gap}c); however such momentum drives FD in the Gradient Cliff (figure \ref{fig:gradient_gap}f) further into the zero-fitness area, highlighting that heuristics like momentum (while useful) do not enable conditional gap-crossing as in high-variance ES, which is informed not by a heuristic, but by empirical evidence of what lies across the gap. Note that a GA's population would also be able to conditionally cross such gaps.

Overall, these landscapes, while simple, help to demonstrate that there are indeed systematic differences between ES and traditional gradient descent.  They also show that no particular treatment is ideal in all cases, so the utility of the optimizing over a fixed-variance search distribution, at least for finding the global optimum, is (as would be expected) domain-dependent. The next section describes results in the Humanoid Locomotion domain that provide a proof-of-concept that these differences also 
manifest themselves when applying ES to modern deep RL benchmarks.

\vspace{-0.05in}
\section{Humanoid Locomotion}



In the Humanoid Locomotion domain, a simulated humanoid robot is controlled by an NN controller with the objective
of producing a fast energy-efficient gait \citep{tassa:mujoco,brockman}, implemented in the Mujoco physics simulator \citep{todorov:mujoco}. Many RL methods are able to produce competent gaits, which this paper considers as achieving a fitness score of 6,000 averaged across many independent evaluations, following the threshold score in \citet{salimans:es}; averaging is necessary because the domain is stochastic. The purpose of this experiment is
not to compare \emph{performance} across methods as is typical in RL, but instead to examine the \emph{robustness} of solutions, as defined by the distribution of performance in the neighborhood of solutions.

Three methods are compared in this experiment: ES, GA, and TRPO. Both ES and GA directly
search through the parameter space for solutions, while TRPO uses gradient descent to modify policies directedly to more often take actions resulting in higher reward. All methods optimize the same underlying NN architecture, which is a feedforward NN with two hidden layers of $256$ Tanh units, comprising approximately 167,000 weight parameters (recall that ES optimizes the same number of parameters, but that they represent the mean of a search distribution over domain parameters). This NN architecture is taken from the configuration file released with the source code from \citet{salimans:es}. The architecture described in their paper is similar, but smaller, having 64 neurons per layer \citep{salimans:es}. 

The hypothesis is that ES policies will be more robust to policy perturbations than policies of similar performance generated by either GA or TRPO. The GA of \citet{such:arxiv17} provides a natural control, because its mutation operator is the same that generates variation within ES, but its objective function does not directly reward robustness. Note that ES is trained with policy perturbations from a Gaussian distribution with $\sigma=0.02$ while the GA required a much narrower distribution ($\sigma=0.00224$) for successful training \citep{such:arxiv17}; 
training the GA with a larger mutational distribution destabilized evolution, as mutation too rarely would preserve or improve performance to support adaptation. Interestingly, this destabilization itself supports the idea that robustness to high variance perturbations is not pervasive throughout this search space. TRPO provides another useful control, because it follows the gradient of increasing performance without generating any random parameter perturbations; thus if the robustness of ES solutions is higher than that of those from TRPO, it also provides evidence that ES's behavior is distinct, i.e.\ it is \emph{not} best understood as simply following the gradient of improving performance with respect to domain parameters (as TRPO does). Note that this argument does not imply that TRPO is deficient if its policies are less robust to random parameter perturbations than ES, as such random perturbations are not part of its search process.

The experimental methodology is to take solutions from different methods and examine the distribution of resulting performance when policies are perturbed with the
perturbation size of ES and of GA. In particular, policies are taken from generation 1,000 of the GA, from iteration 100 of ES, and from 
iteration 350 of TRPO, where methods have approximately evolved a solution of $\approx$6,000 fitness. 
The ES is run with hyperparameters according to \citet{salimans:es}, the GA is taken from \citet{such:arxiv17}, and
TRPO is based on OpenAI's baselines package \citep{baselines}. Exact hyperparameters are listed in the supplemental material.


\vspace{-0.075in}
\subsection{Results}

\begin{figure}
  \centering
    \begin{subfigure}[t]{0.5\textwidth}
  		\centering
         \includegraphics[height=1.6in]{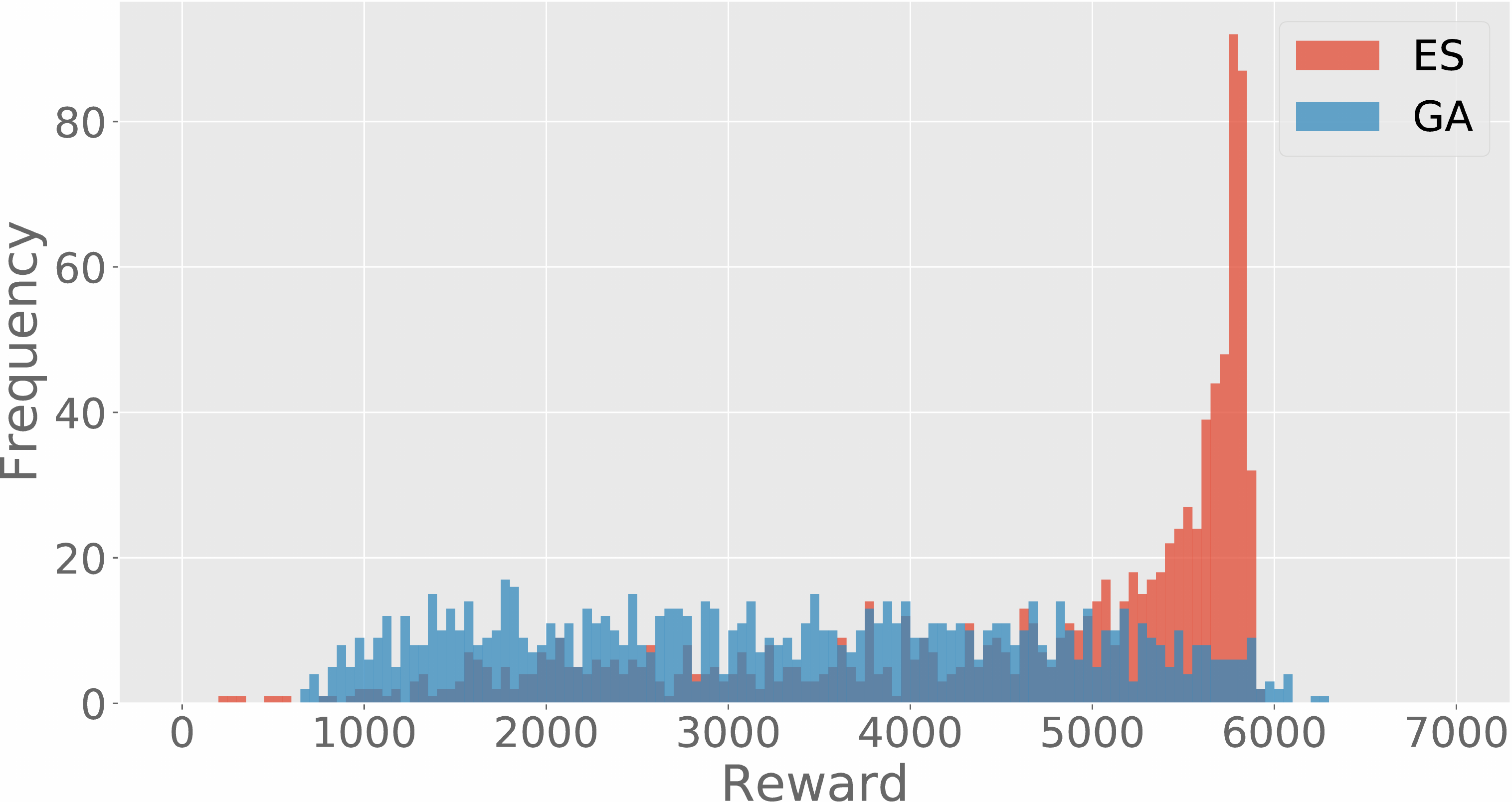}
	    \vspace{-0.02in}
        \caption{ES ($\sigma=0.02$) vs GA ($\sigma=0.002$)}
    \end{subfigure}%
	\\ \vspace{0.1in}
    \begin{subfigure}[t]{0.5\textwidth}
  		\centering
         \includegraphics[height=1.6in]{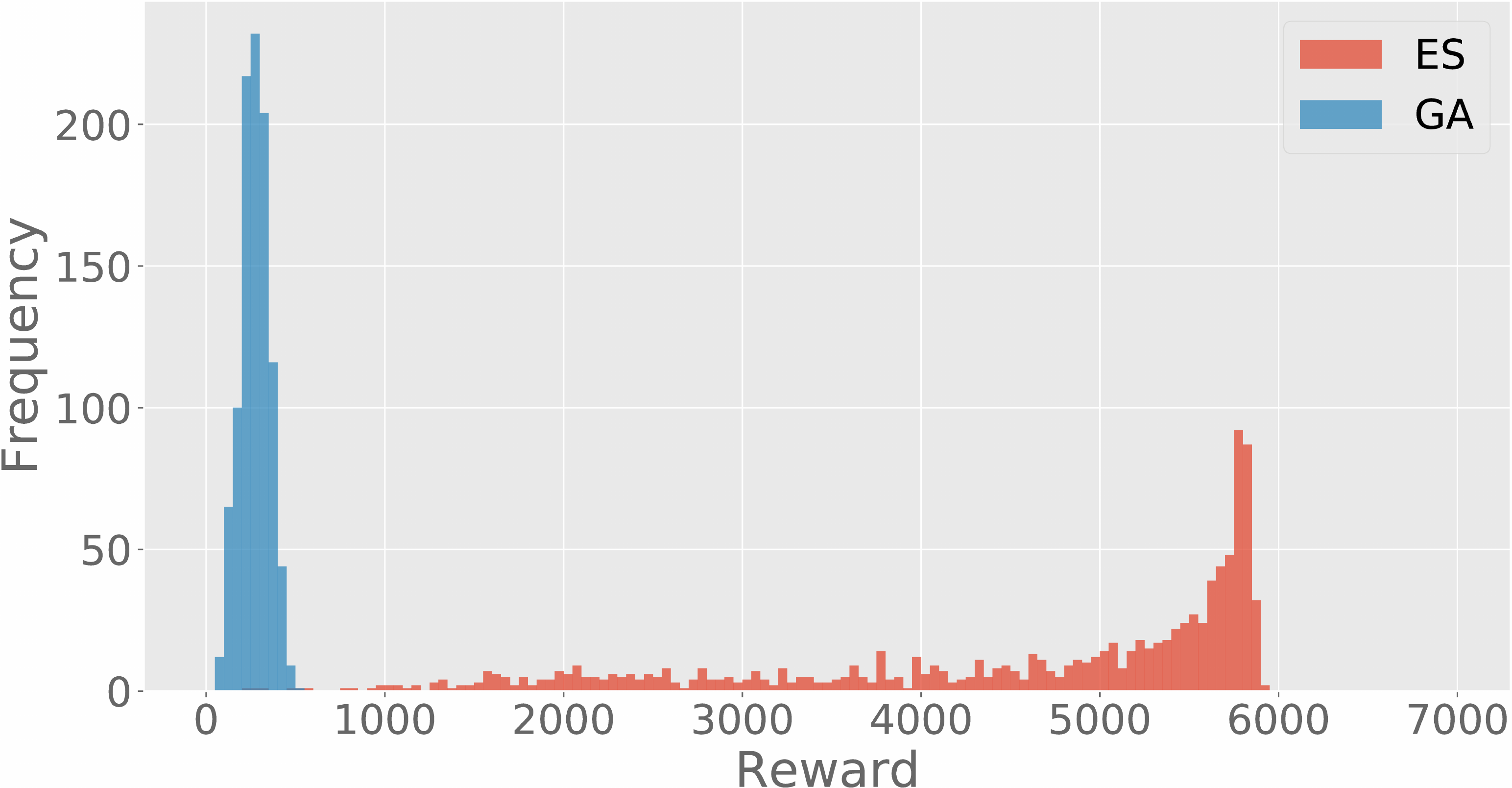}
	    \vspace{-0.02in}
        \caption{ES ($\sigma=0.02$) vs GA ($\sigma=0.02$)}
    \end{subfigure}    
	\\ \vspace{0.1in}
     \begin{subfigure}[t]{0.5\textwidth}
  		\centering
         \includegraphics[height=1.6in]{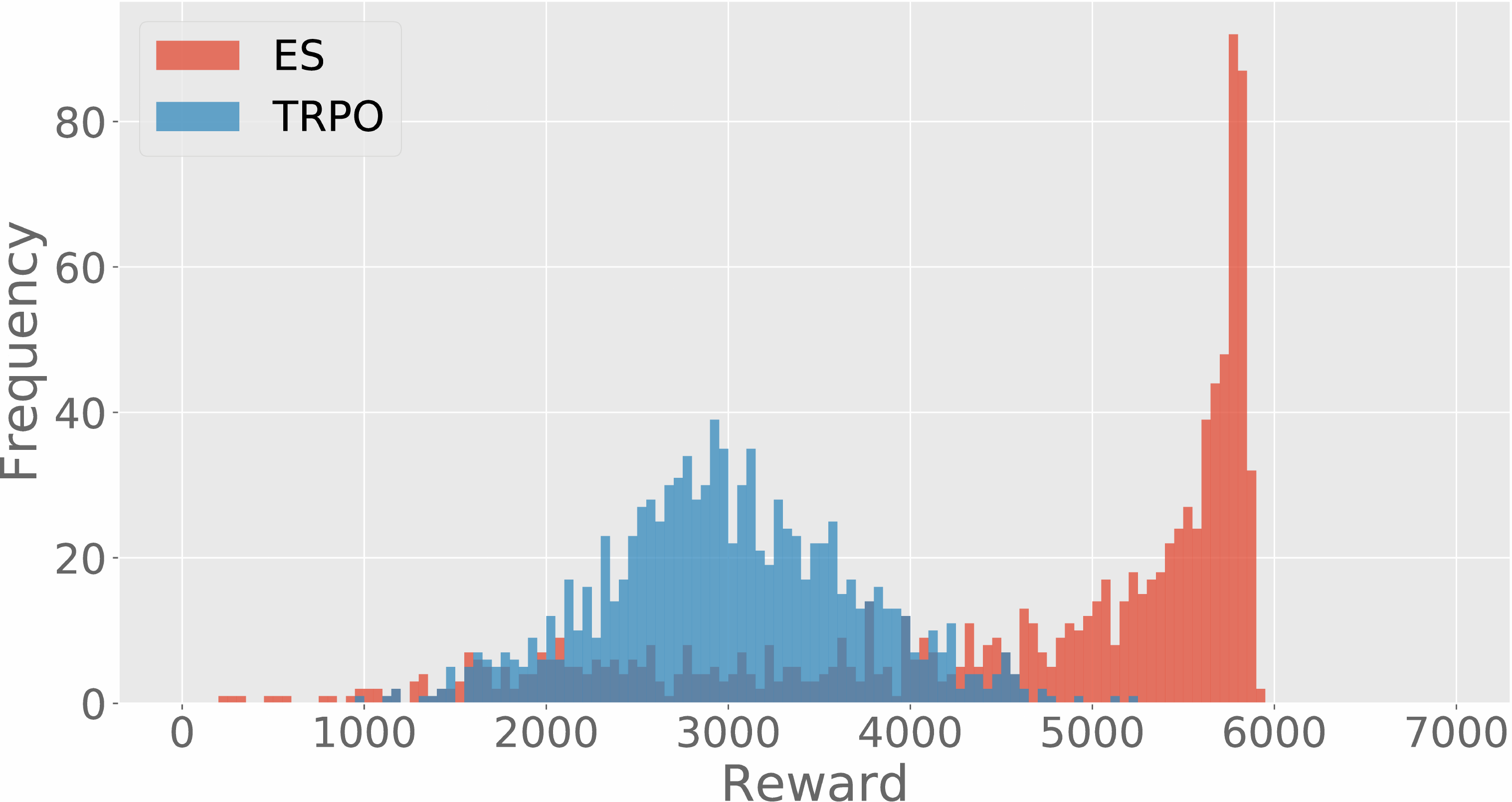}
	    \vspace{-0.02in}
        \caption{ES ($\sigma=0.02$) vs TRPO ($\sigma=0.02$)}
    \end{subfigure}%
	\vspace{-0.1in}
\caption{\textbf{ES is more robust to parameter perturbation in the Humanoid Locomotion task.} The distribution of reward is shown from perturbing models trained by ES, GA, and TRPO. Models were trained to a fitness value of 6,000 reward, and robustness is evaluated by generating perturbations with Gaussian noise (with the specified variance) and evaluating perturbed policies in the domain. High-variance perturbations of ES produce a healthier distribution of reward than do perturbations of GA or TRPO. \label{fig:es_vs_ga_resiliency}}
	\vspace{-0.1in}
\end{figure}

Figure \ref{fig:es_vs_ga_resiliency} shows a representative example of the stark difference between the robustness of ES solutions and those from the GA or TRPO, even when the GA is subject only to the lower-variance perturbations that were applied during evolution. We observed that this result appears consistent across independently trained models. A video comparing perturbed policies of ES and TRPO can be viewed at the following URL (along with other videos showing selected fitness landscape animations): \url{https://goo.gl/yz1MeM}. Note that future work could explore whether modifications to the vanilla GA would result in similar robustness as ES \cite{lehman:selfadapt,meyer:selfadapt}.

\begin{figure}
  		\centering
         \includegraphics[height=1.7in]{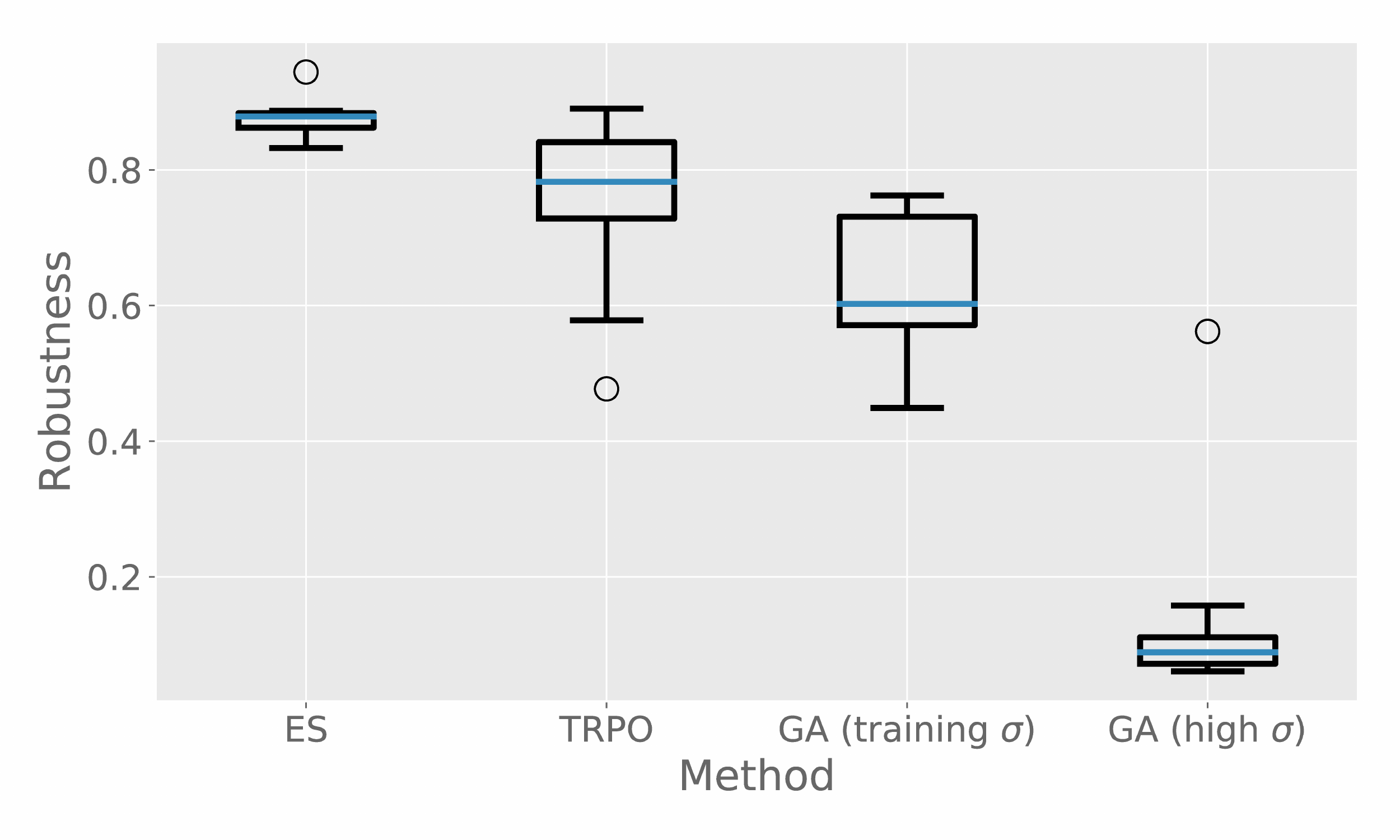}
\vspace{-0.15in}
\caption{\textbf{Quantitative measure of robustness across independent runs of ES, GA, and TRPO.} The distribution of reward is shown from perturbing ten independent models for each of ES, GA, and TRPO under the high-variance perturbations used to train ES ($\sigma=0.02$). Results from GA are shown also for perturbations drawn from the lower-variance distribution it experienced during training ($\sigma=0.00224$). The conclusion is that high-variance perturbations of ES retain significantly higher performance than do perturbations of GA or TRPO (Mann-Whitney U-test; $p<0.01$). \label{fig:humanoid_boxplot}}
	\vspace{-0.15in}
\end{figure}

To further explore this robustness difference, a quantitative measure of robustness was also applied.
In particular, for each model, the original parameter vector's reward was calculated by averaging its performance over 1,000 trials in the environment. Then, 1,000 perturbations were generated for each model, and each perturbation's performance was averaged over 100 trials in the environment. Finally, a robustness score is calculated for each model as the ratio of the perturbations' median performance to the unperturbed policy's performance, i.e.\ a robustness score of 0.5 indicates that the median perturbation performs half as well as the unperturbed model. The results (shown in figure \ref{fig:humanoid_boxplot}) indicate that indeed by this measure ES is significantly more robust than the GA or TRPO (Mann-Whitney U-test; $p<0.01$). The conclusion is that the robustness-seeking property of ES demonstrated in the simple landscapes also manifests itself in this more challenging and high-dimensional domain. Interestingly, TRPO is significantly more robust than both GA treatments (Mann-Whitney U-test; $p<0.01$) even though it is not driven by random perturbations; future work could probe the relationship between the SGD updates of policy gradient methods and the random perturbations applied by ES and the GA.  

\vspace{-0.125in}
\section{Discussion and Conclusion}


An important contribution of this paper is to ensure that awareness of the robustness-seeking property of ES, especially with higher $\sigma$, is not lost -- which is a risk when ES is described as simply performing stochastic finite differences.  When $\sigma$ is above some threshold, it is not accurate to interpret ES as merely an approximation of SGD, nor as a traditional FD-based approximator.  Rather, it becomes a gradient approximator coupled with a compass that seeks areas of the search space robust to parameter perturbations.  This latter property is not easily available to point-based gradient methods, as highlighted dramatically in the Humanoid Locomotion experiments in this paper.  On the other hand, if one wants ES to better mimic FD and SGD, that option is still feasible simply by reducing $\sigma$.

The extent to which seeking robustness to parameter perturbation is important remains open to further research.  As shown in the landscape experiments, when it comes to finding optima, it clearly depends on the domain.  If the search space is reminiscent of Fleeting Peaks, then ES is likely an attractive option  for reaching the global optimum. However, if it is more like the Narrowing Path landscape, especially if the ultimate goal is a single solution (and there is no concern about its robustness), then high-sigma ES is less attractive (and the lower-sigma ES explored in \citet{zhang:arxiv17} would be more appropriate). It would be interesting to better understand whether and under what conditions domains more often resemble Fleeting Peaks as opposed to the Narrowing Path.  

An intriguing question that remains open is when and why such robustness might be desirable even for reasons outside of global optimality.  For example, it is possible that policies encoded by networks in robust regions of the search space (i.e.\ where perturbing parameters leads to networks of similar performance) are also robust to other kinds of noise, such as domain noise.  It is interesting to speculate on this possibility, but at present it remains a  topic for future investigation.  Perhaps parameter robustness also correlates to robustness to new opponents in coevolution or self-play, 
but that again cannot yet be answered.  
Another open question is how robustness interacts with divergent search techniques like novelty search \citep{lehman:ecj11} or quality diversity methods \citep{pugh:frontiers16}; follow-up experiments to \citet{conti:arxiv17}, which combines ES with novelty search, could explore this issue.
Of course, the degree to which the implications of robustness matter likely varies by domain as well.  For example, in the Humanoid Locomotion task the level of domain noise means that there is little choice but to choose a higher $\sigma$ during evolution (because otherwise the effects of perturbations could be drowned out by noise), but in a domain like MNIST there is no obvious need for anything but an SGD-like process \citep{zhang:arxiv17}.

Another benefit of robustness is that it could indicate \emph{compressibility}: If small mutations tend not to impact functionality (as is the case for robust NNs), then less numerical precision is required to specify an effective set of network weights (i.e.\ fewer bits are required to encode them). This issue too is presently unexplored.

This study focused on ES, but it raises new questions about other related algorithms. For instance, non-evolutionary methods may be modified to include a drive towards robustness or may already share abstract connections with ES. For example, stochastic gradient Langevin dynamics \citep{welling:sgd}, a Bayesian approach to SGD, approximates a distribution of solutions over iterations of training by adding Gaussian noise to SGD updates, in effect also producing a solution cloud. Additionally, it is possible that methods combining parameter-space exploration with policy gradients (such as \citeauthor{plappert:parameter} \citep{plappert:parameter}) could be modified to include robustness pressure.

A related question is, do all population-based EAs possess at least the potential for the same tendency towards robustness \citep{wilke:nature01,wagner:robustness,lenski:balancing}? Perhaps some such algorithms have a different means of turning the knob between gradient following and robustness seeking, but nevertheless in effect leave room for the same dual tendencies. One particularly interesting relative of ES is the NES \citep{wierstra:jmlr14}, which adjusts $\sigma$ dynamically over the run.  Given that $\sigma$ seems instrumental in the extent to which robustness becomes paramount, characterizing the tendency of NES in this respect is also important future work. 

We hope ultimately that the brief demonstration in this work can serve as a reminder that the analogy between ES and FD only goes so far, and there are therefore other intriguing properties of ES that remain to be investigated.




\vspace{-0.1in}
\section*{Acknowledgements}
We thank the members of Uber AI Labs, in particular Thomas Miconi, Martin Jankowiak, Rui Wang, Xingwen Zhang, and Zoubin Ghahramani for helpful discussions; Felipe Such for his GA implementation and Edoardo Conti for his ES implementation. We also thank Justin Pinkul, Mike Deats, Cody Yancey, Joel Snow, Leon Rosenshein and the entire OpusStack Team inside Uber for providing our computing platform and for technical support.


\vspace{-0.12in}
\bibliography{es.bib,nn,ucf}
\bibliographystyle{apalike}

%% file: si.tex
\end{comment}
\pagebreak

\section*{Supplemental Material: Hyperparameters}

This section describes the relevant hyperparameters for the search methods (ES, GA, and TRPO) applied in the Humanoid Walker experiments.

\subsection*{ES}

The ES algorithm was based on \citet{salimans:es} and uses the same hyperparameters as in their Humanoid Walker experiment. In particular, 10,000 domain roll-outs were used per iteration of the algorithm, with a fixed $\sigma$ of the parameter distribution set to 0.02. Note that the $\sigma$ hyperparameter was not selected to maximize robustness (i.e.\ it was taken from \citet{salimans:es}, and it was chosen there for performance reasons). The ADAM
optimizer was applied with a step-size of 0.01. 

Analyzed champions were taken from iteration 100 of ES, i.e.\ a total of 1$,$000$,$000 domain roll-outs were expended to reach that point in the search.

\subsection*{GA}

The GA was based on \citet{such:arxiv17}. The population size was set to 12,501, and $\sigma$ of the normal distribution used to generate mutation perturbations was set to 0.00224. Truncation selection was performed, and only the highest-performing 5\% of the population survived. The fitness score for each individual was the average of five noisy domain roll-outs.

Analyzed GA champions were taken from generation 1,000 of the GA, which means a total of 62,505,000 domain evaluations were expended to reach that point. For each run, the 20 highest-fitness individuals in the population were each evaluated in 100 additional domain roll-outs, and the individual with average performance closest to 6,000 was the one selected for further analysis.

\subsection*{TRPO}

The TRPO \citep{schulman:trpo} implementation was taken from the OpenAI baselines package \citep{baselines}. The maximum KL divergence was set to 0.1 and 10 iterations of conjugate gradients were conducted per batch of training data. Discount rate ($\gamma$) was set to 0.99. 

Analyzed policies were taken from independent runs at iteration 350, and each run was parallelized across 60 worker threads. Each iteration consumes 1,024 simulation steps for each worker thread, thus requiring a total of 21,504,000 simulation steps. While it is difficult to make a direct comparison between the number of simulation steps (for TRPO) and complete domain rollouts (i.e.\ episodes run from beginning to end, for GA and ES),  on a gross level
TRPO did require less simulation computation than either ES or GA (i.e.\ TRPO was more sample efficient in this domain).

